\documentclass[journal]{IEEEtran}
\usepackage{amsmath,amsfonts}
\usepackage{amsthm}
\usepackage{algorithmic}
\usepackage{algorithm}
\usepackage{array}
\usepackage[caption=false,font=normalsize,labelfont=sf,textfont=sf]{subfig}
\usepackage{textcomp}
\usepackage{stfloats}
\usepackage{float}
\usepackage[breakable]{tcolorbox}
\usepackage[table]{xcolor}
\usepackage{url}
\usepackage{verbatim}
\usepackage{graphicx}
\usepackage{multirow}
\usepackage{multicol}
\usepackage{cite}
\usepackage{balance}

\usepackage{hyperref} 
\hypersetup{
    colorlinks=true,
    linkcolor=blue,
    citecolor=blue,
    urlcolor=teal
}

\definecolor{myhighlight}{rgb}{0.77, 0.76, 0.82}
\definecolor{darkgreen}{rgb}{0.0, 0.5, 0.0}

\begin{document}

\title{Reservoir-Based Graph Convolutional Networks}

\author{Mayssa Soussia\textsuperscript{1,2\dag},~Gita Ayu Salsabila\textsuperscript{2\dag},~Mohamed Ali Mahjoub\textsuperscript{1}~and~Islem Rekik\textsuperscript{2*}
\thanks{\textsuperscript{\dag}Mayssa Soussia and Gita Ayu Salsabila are co-first authors.}
\thanks{\textsuperscript{*}Corresponding author: i.rekik@imperial.ac.uk, \url{https://basira-lab.com/}}
\thanks{\textsuperscript{1}National Engineering School of Sousse, University of Sousse, LATIS -- Laboratory of Advanced Technology and Intelligent Systems, 4023, Sousse, Tunisia.}
\thanks{\textsuperscript{2}BASIRA Lab, Imperial-X and Department of Computing, Imperial College London, London, UK.}}

\markboth{IEEE Transactions on Pattern Analysis and Machine Intelligence}%
{Soussia \MakeLowercase{\textit{et al.}}: Reservoir-Based Graph Convolutional Networks}

\maketitle

\begin{abstract}
Message passing is a core mechanism in Graph Neural Networks (GNNs), enabling the iterative update of node embeddings by aggregating information from neighboring nodes. Graph Convolutional Networks (GCNs) exemplify this approach by adapting convolutional operations for graph structures, allowing features from adjacent nodes to be combined effectively. However, GCNs encounter challenges with complex or dynamic data. Capturing long-range dependencies often requires deeper layers, which not only increase computational costs but also lead to over-smoothing, where node embeddings become indistinguishable. To overcome these challenges, reservoir computing has been integrated into GNNs, leveraging iterative message-passing dynamics for stable information propagation without extensive parameter tuning. Despite its promise, existing reservoir-based models lack structured convolutional mechanisms, limiting their ability to accurately aggregate multi-hop neighborhood information. To address these limitations, we propose RGC-Net (\emph{Reservoir-based Graph Convolutional Network}), which integrates reservoir dynamics with structured graph convolution. Key contributions include: (i) a reimagined convolutional framework with fixed-random reservoir weights and a leaky integrator to enhance feature retention; (ii) a robust, adaptable model for graph classification; and (iii) an RGC-Net-powered transformer for graph generation with application to dynamic brain connectivity. Extensive experiments show RGC-Net achieves state-of-the-art performance in classification and generative tasks, including brain graph evolution, with faster convergence and mitigated over-smoothing. Our source code is available at \url{https://github.com/basiralab/RGC-Net}.
\end{abstract}


\section{Introduction}
\IEEEPARstart{G}{raph} neural networks (GNNs) have opened new pathways in machine learning, mainly designed to operate on graph-structured data \cite{wu2020comprehensive,zhou2020graph}. By capturing patterns across networks of interconnected entities---whether molecules \cite{li2021graph}, social ties \cite{sharma2024survey}, or brain regions \cite{bessadok2022graph}---GNNs allow us to understand and predict outcomes that hinge on both individual characteristics and their links. The core learning operation in GNNs is the message-passing mechanism, which recursively updates each node's embedding by aggregating information from neighboring nodes along with its own features \cite{zhou2020graph,wu2022graph}. This operation facilitates the propagation of information across the graph, allowing each node to gain knowledge about its local neighborhood and the graph global structure.

Numerous GNN variants have emerged, each distinguished by a unique message-passing mechanism, such as the Graph Attention Network (GAT) \cite{velivckovic2017graph}, Graph Isomorphism Network (GIN) \cite{xu2018powerful}, and Graph Autoencoders \cite{kipf2016variational}. However, Graph Convolutional Networks (GCNs) \cite{kipf2016semi} have garnered the lion's share of attention \cite{zhang2019graph, bhatti2023deep}. Inspired by the convolutional principles applied to grid-based image data, GCNs adapt this approach to accommodate the irregular structure of graphs, thereby extending convolutional techniques to capture and leverage the complex, relational nature of interconnected data. Specifically, \emph{message passing} in GCNs is performed through matrix multiplication involving the graph connectivity matrix, the node feature matrix, and a weight matrix that transforms node features \cite{kipf2016semi}.

GCNs have proven their versatility and effectiveness across diverse applications \cite{bhatti2023deep} thanks to their ability to harness information from neighboring nodes in an efficient and scalable manner. However, while GCNs are well-suited to tasks requiring static graph representations, they encounter significant challenges when dealing with complex, dynamic, or sequential data \cite{li2018deeper}. This limitation is further exacerbated by the need for deeper layers to capture information from more distant nodes; as the receptive field expands, so does the computational complexity, resulting in higher training costs due to the increasing number of parameters \cite{li2018deeper}. Furthermore, deeper GCNs are prone to over-smoothing, where node embeddings converge to similar values, ultimately diminishing their ability to represent distinct node characteristics effectively across the graph \cite{yang2020revisiting}.

These challenges have led to the exploration of alternative architectures, such as reservoir-based GNNs \cite{Gallicchio2010GraphEcho,gallicchio2019fastdeepgraphneural,tortorella2022leavegraphsaloneaddressing,BIANCHI2022389,pasa2021multiresolution}, which incorporate fixed dynamic systems within the message-passing framework, aiming to overcome the limitations of conventional GCNs. By leveraging a fixed reservoir, these models retain richer feature representations over extended iterations, enabling efficient information propagation across the graph without increasing the parameter count. This approach reduces computational complexity while improving representation quality by preserving distinct node features and capturing long-range dependencies. Results show that reservoir computing in GNNs achieves comparable or superior performance in graph classification tasks \cite{BIANCHI2022389,pasa2021multiresolution} while also offering faster training \cite{Gallicchio2010GraphEcho}.

While reservoir-based GNNs, including GraphESN \cite{Gallicchio2010GraphEcho}, have shown promise, they also introduce certain constraints. \emph{First}, the contractive dynamics inherent to these models enforce a Markovian process, which biases them toward short-term dependencies and limits their effectiveness in tasks requiring long-range dependency modeling across the graph structure. \emph{Second}, the fixed reservoir structure in these models results in inefficient multi-hop aggregation, reducing precision in capturing intricate graph relationships as the architecture lacks optimization for layer-wise or hierarchical neighborhood aggregation. \emph{Third}, the application scope of many reservoir-based GNNs remains narrow, often confined to static graph classification tasks, which limits their adaptability to dynamic or temporal graph applications. These applications, such as brain graph evolution prediction, focus on developing graph generative models capable of capturing and forecasting the temporal dynamics of brain connectivity at different stages or time points \cite{gurler2020foreseeing, nebli2020deep, tekin2021recurrent, demirbilek2023predicting}.

In response to these gaps, we propose RGC-Net (Reservoir-based Graph Convolutional Network), a novel architecture that redefines reservoir dynamics by explicitly integrating the principles of graph convolution into its update rule. Unlike prior models \cite{Gallicchio2010GraphEcho,gallicchio2019fastdeepgraphneural} that rely on generic, non-interpretable state transitions, RGC-Net introduces a \emph{layered graph convolutional mechanism within the reservoir framework}. To further enhance RGC-Net capacity for preserving meaningful features across iterations, we incorporate a leaky integrator within the reservoir. This mechanism balances retained past information with new data, mitigating over-smoothing effects without relying solely on the contraction properties typical of other reservoir-based GNNs. Such integration allows RGC-Net to maintain the expressiveness of node representations over multiple layers, enabling more robust graph learning for deep architectures and dynamic graph data. Furthermore, by introducing a trainable counterpart, TRGC-Net, our framework allows for the first systematic comparison between fixed and trainable reservoir weights, offering new insights into the role of non-trainable reservoirs in graph representation learning.

The key contributions of our work are outlined as follows:
\begin{enumerate}
    \item \textbf{RGC-Net Architecture}: we introduce RGC-Net (Reservoir-based Graph Convolutional Network), a hybrid model that synergizes the structured, multi-hop neighborhood aggregation of graph convolution with the dynamic, recursive characteristics of reservoir computing. By integrating a reservoir with fixed, non-trainable parameters and leaky integration into the graph convolutional framework, RGC-Net achieves a balance: it retains GCN's structured aggregation while leveraging reservoir dynamics to mitigate over-smoothing and reduce training costs. The RGC-Net architecture is built upon three core hypotheses:

    \begin{tcolorbox}[colback=gray!10, colframe=black, boxrule=0.2pt, breakable]
        \textbf{\textit{Hypothesis 1.}} \textit{The dynamic and recursive nature of reservoir computing can perform graph convolution as effectively or better than traditional graph convolution.}

        \vspace{0.2cm}
        \textbf{\textit{Hypothesis 2.}} \textit{The leaky integrator of the reservoir can control how much of the initial node embedding is retained during node aggregation, preserving the original node characteristics and preventing over-smoothing.}

        \vspace{0.2cm}
        \textbf{\textit{Hypothesis 3.}} \textit{Non-trainable parameters in the reservoir can lead to faster convergence and lower resource consumption during model training.}
    \end{tcolorbox}

    \item \textbf{Broad applicability in graph learning}: we apply RGC-Net to both graph classification and brain graph evolution prediction tasks. This includes its role as a robust graph classifier across diverse datasets and its integration into a graph transformer framework for temporal brain graph generation, supporting future brain connectivity prediction based on current data.

    \item \textbf{Comprehensive evaluation and analysis}: we conduct an extensive evaluation of RGC-Net across various graph learning tasks, assessing its performance, efficiency, and adaptability. Our experiments explore the effects of hyperparameters such as layer depth, iterations, and leaky rate, highlighting RGC-Net's computational efficiency and scalability advantages over conventional models.
\end{enumerate}

\section{Related Work}
\subsection{Reservoir Computing in GNNs}

Early Graph Neural Networks (GNNs) integrated Recurrent Neural Networks (RNNs) to iteratively learn node representations, achieving robust latent embeddings for graph nodes through recurrent updates \cite{sperduti1997supervised, micheli2004contextual, scarselli2008graph}. This process iteratively propagates node information until convergence, contracting node embeddings in latent space and ensuring comprehensive information flow across the graph. Reservoir computing, a specialized RNN variant with non-trainable, fixed random weights, has emerged as an efficient approach for embedding updates in GNNs, offering reduced computational costs by bypassing backpropagation on the recurrent structure \cite{Gallicchio2010GraphEcho, gallicchio2019fastdeepgraphneural, tortorella2022leavegraphsaloneaddressing, BIANCHI2022389, pasa2021multiresolution}. By initializing the reservoir with fixed weights, these models ensure contractive dynamics, typically setting the spectral radius of the weight matrix below one to maintain stability. Gallicchio and Micheli \cite{Gallicchio2010GraphEcho} demonstrated that this spectral adjustment enables any reservoir to converge to an equilibrium state, thereby preserving stable, consistent dynamics across iterative updates while promoting efficient information diffusion through the graph structure.

GraphESN \cite{Gallicchio2010GraphEcho} pioneered reservoir-based state transitions for iteratively updating node embeddings, using fixed weights for target nodes and their neighbors to efficiently capture structural features without additional training overhead. Building on this foundation, FDGNN \cite{gallicchio2019fastdeepgraphneural} introduced a stacked architecture of reservoir layers, enhancing classification accuracy and reducing training times by enabling deeper feature extraction. Bianchi et al. \cite{BIANCHI2022389} advanced this approach by adding pooling layers after each reservoir layer, thereby reducing embedding dimensionality and improving computational efficiency, which made it more suitable for larger graph datasets. Despite these advances, over-smoothing remains a challenge in these architectures, as repeated iterations cause node embeddings to converge, losing distinctiveness across nodes \cite{zhou2020graph}. Addressing this, Pasa et al. proposed the Multiresolution Reservoir GNN (MRGNN) \cite{pasa2021multiresolution}, which limits message passing to a specific neighborhood radius, preserving local node diversity and enhancing stability, especially beneficial for tasks sensitive to local node variations. However, while this approach addresses over-smoothing in local neighborhoods, it, along with other reservoir-based models, remains limited in its ability to capture long-range dependencies and dynamic graph applications, highlighting the need for more flexible, scalable solutions in reservoir-based GNNs.

\subsection{Brain Graph Evolution Prediction}
Researchers have explored various methods to predict evolving brain connectivity in brain graphs over time. Early approaches employed supervised sample selection techniques, averaging brain graphs of similar subjects in the training set to generate follow-up graphs. For instance, Ghribi et al. \cite{ghribi2021multi} used bidirectional regressors for similarity scoring between input and training subjects, while Ezzine and Rekik \cite{ezzine2019learning} proposed the LINAs framework, ranking representative brain atlases in an unsupervised manner to guide sample selection. However, such sample selection-based methods require substantial computational resources due to complex selection and averaging processes, particularly with large graph datasets.

More recent studies have leveraged message-passing Graph Neural Networks (MP-GNNs) for brain evolution prediction, learning brain graph representations over time. Hong et al. \cite{hong2019longitudinal} developed an adversarial network with GCN layers to predict missing longitudinal brain data, mapping graphs across time points to construct graphs at missing intervals. Similarly, G\"okta\c{s} et al. \cite{goktacs2020residual} introduced RESNets, a GCN-based autoencoder that utilizes a Connectional Brain Template (CBT) to guide follow-up brain graph predictions. Other approaches, like gGAN \cite{gurler2020foreseeing} and EvoGraphNet \cite{nebli2020deep}, incorporate adversarial frameworks for brain graph generation, using GCN-based discriminators and generators. While these MP-GNN models capture brain graph topology, their reliance on GCNs poses challenges, as deeper layers required for long-range dependencies lead to over-smoothing and scaling issues in high-resolution graphs.

To better capture temporal dynamics, recurrent graph models like the Recurrent Brain Graph Mapper (RBGM) \cite{tekin2021recurrent} and ReMI-Net \cite{demirbilek2023predicting} have been proposed. RBGM leverages recurrent graph networks to map brain evolution across time points, while ReMI-Net identifies distinctive biomarkers in healthy versus disordered populations by learning population-level brain templates over time. Despite promising results, these recurrent models face limitations with memory-intensive, edge-based convolution, posing challenges for high-resolution brain graphs.

In summary, while existing methods offer valuable insights, most studies lack evaluation on high-resolution brain data. Current MP-GNN and recurrent graph models remain largely untested for scalability and generalizability with fine-grained brain graphs, leaving an open challenge for robust, high-resolution brain graph evolution prediction.

\section{Background and Preliminaries}
\subsection{Echo State Network (ESN)}
\begin{figure}[!t]
    \centering
    \includegraphics[width=1\linewidth]{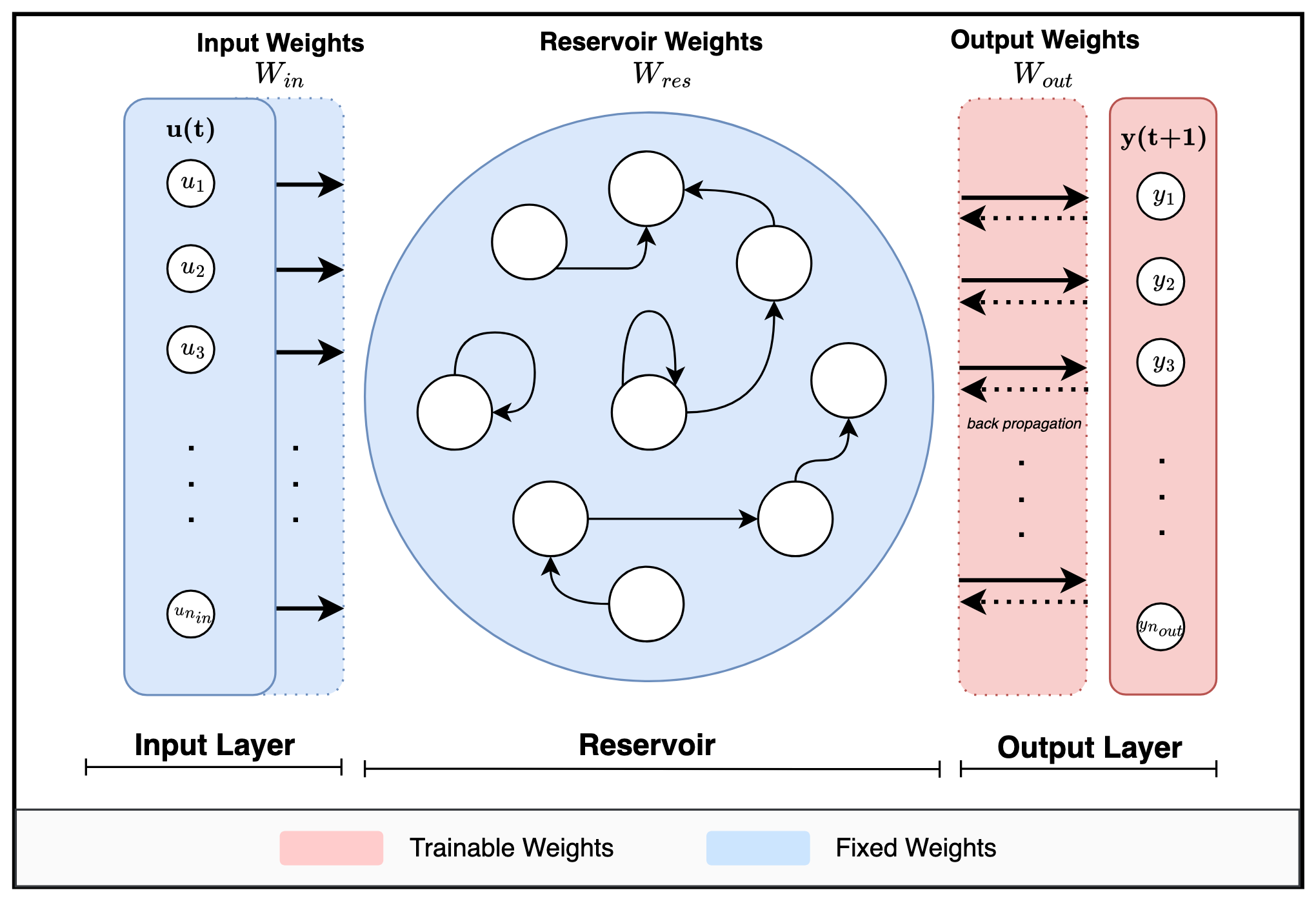}
    \caption{\textit{Echo State Network (ESN) architecture with an input layer, a fixed-weight reservoir, and an output layer. Input weights \( W_{\text{in}} \in \mathbb{R}^{n_{\text{res}} \times n_{\text{in}}} \) map the input vector \( u(t) \in \mathbb{R}^{n_{\text{in}}} \) into the reservoir. The reservoir uses recurrent weights \( W_{\text{res}} \in \mathbb{R}^{n_{\text{res}} \times n_{\text{res}}} \) to capture temporal dependencies. The output layer, with weights \( W_{\text{out}} \in \mathbb{R}^{n_{\text{out}} \times n_{\text{res}}} \), generates the output vector \( y(t+1) \in \mathbb{R}^{n_{\text{out}}} \), with only \( W_{\text{out}} \) being trainable.}}
    \label{fig:ESN}
\end{figure}

An Echo State Network (ESN) is a specialized recurrent neural network (RNN) architecture designed to efficiently process sequential and time-dependent data. Initially developed for sequence prediction tasks, ESNs have demonstrated broad applicability across domains such as image recognition \cite{tong2018reservoir}, reinforcement learning \cite{chang2019convolutional}, and graph classification \cite{BIANCHI2022389,pasa2021multiresolution}. Conventionally, an ESN leverages a fixed \textit{reservoir}---a large, sparsely connected layer of hidden neurons with random weights---to project input data into high-dimensional representations. The ESN operates according to the following equations:
\begin{equation}
    x(t+1) = \sigma(W_{\text{in}} u(t) + W_{\text{res}} x(t))
    \label{eq:state_update}
\end{equation}
\begin{equation}
    y(t+1) = W_{\text{out}} x(t+1)
    \label{eq:output}
\end{equation}
where \( u(t) \in \mathbb{R}^{n_{\text{in}}} \) is the input vector at time \( t \), \( x(t) \in \mathbb{R}^{n_{\text{res}}} \) represents the hidden state within the reservoir, and \( y(t+1) \in \mathbb{R}^{n_{\text{out}}} \) is the output vector. \( W_{\text{in}} \in \mathbb{R}^{n_{\text{res}} \times n_{\text{in}}} \) maps the input \( u(t) \) to the reservoir state space, \( W_{\text{res}} \in \mathbb{R}^{n_{\text{res}} \times n_{\text{res}}} \) defines the internal connections within the reservoir, consisting of fixed random weights, and \( W_{\text{out}} \in \mathbb{R}^{n_{\text{out}} \times n_{\text{res}}} \) projects the high-dimensional reservoir states to the output. Only \( W_{\text{out}} \) is optimized during training, allowing for computational efficiency. For stability, the \textit{spectral radius} of \( W_{\text{res}} \) (the largest absolute eigenvalue) is typically set below 1, enforcing a contraction mapping that stabilizes the reservoir dynamics over time \cite{jaeger2001echo}. Additionally, a \textit{leaky integrator} modulates memory retention, balancing past and new information in the hidden state updates. The ESN architecture is illustrated in Figure \ref{fig:ESN}.

\subsubsection{Reservoir Spectral Radius Adjustment}
The spectral radius of the reservoir, \( \rho(\mathbf{W}_{\text{res}}) \), represents the largest absolute eigenvalue of the reservoir weight matrix and serves as a fundamental stability criterion. For the reservoir to maintain the Echo State Property, it is essential that \( \rho(\mathbf{W}_{\text{res}}) < 1 \); this ensures that any internal state dynamics induced by input signals will gradually decay, avoiding explosive growth or vanishing effects over time \cite{jaeger2001echo}. If \( \rho(\mathbf{W}_{\text{res}}) \) exceeds 1, the reservoir internal states can grow uncontrollably, leading to an excessive sensitivity to noise. Conversely, if \( \rho(\mathbf{W}_{\text{res}}) \) is too small, the state dynamics may decay too rapidly, hindering the network's ability to capture long-term dependencies and temporal patterns. This adjustment is performed only once, immediately after random initialization, and before any training or inference begins. To guarantee this stability, it is often necessary to normalize the reservoir's weight matrix by scaling it according to its spectral radius if it exceeds 1, as outlined in Algorithm \ref{alg:spectral_radius_adjustment}.

\begin{algorithm}[!t]
\caption{Reservoir Weights Spectral Radius Adjustment}\label{alg:spectral_radius_adjustment}
\begin{algorithmic}
\STATE \textbf{Input:} Reservoir weight matrix \( \mathbf{W} \)
\STATE \textbf{Output:} Adjusted reservoir weight matrix \( \mathbf{W}_{\text{adj}} \)
\STATE \textbf{Compute} the eigenvalues \( \lambda_i \) of \( \mathbf{W} \)
\STATE \textbf{Compute} the spectral radius \( \rho(\mathbf{W}) \) of \( \mathbf{W} \)
\IF{ \( \rho(\mathbf{W}) > 1 \) }
    \STATE \textbf{Compute} the scaling factor (\( \text{scaling\_factor} = \frac{1}{\rho(\mathbf{W})} \))
    \STATE \textbf{Adjust} the weight matrix (\( \mathbf{W}_{\text{adj}} = \text{scaling\_factor} \times \mathbf{W} \))
\ENDIF
\STATE \textbf{Return} \( \mathbf{W}_{\text{adj}} \)
\end{algorithmic}
\end{algorithm}

\subsubsection{Leaky Integrator}
The Echo State Network (ESN) model often utilizes a leaky integrator neuron to process input over time while ``leaking'' or decaying past activity \cite{lun2015novel}. This leaky integrator enables the ESN reservoir to balance between retaining and forgetting past inputs, allowing it to selectively store and process temporal information. The leaky integrator is implemented in ESN as follows:

\begin{equation}
    x(t + 1) = (1 - \alpha) x(t) + \alpha \sigma(W_{\text{in}} u(t) + W_{\text{res}} x(t))
    \label{eq:leaky_integrator_esn}
\end{equation}

where \( \alpha \) denotes the leaky rate. The leaky rate \( \alpha \) is a crucial parameter that controls the rate at which the integrator leaks or decays its previous state. Specifically, it determines the proportion of the previous state retained versus updated based on new input. A larger \( \alpha \) increases the influence of the new input on the updated state, reducing the retention of information from previous states.

\subsection{Graph Neural Networks (GNNs)}
Graph neural networks (GNNs) learn graph representations by capturing both topological structures and node features through iterative message passing, which models information diffusion across the network \cite{kipf2016semi}. Two core components in GNNs, message passing and graph convolution, facilitate this process.

\subsubsection{Message passing}
Message passing is a core operation in GNNs. It updates the representation (or embedding) of each node in the input graph by exchanging information with its neighboring nodes. This process is repeated over multiple layers or iterations, allowing each node to gather information from neighbors multiple hops away. For example, to propagate information from neighbors two hops away, two message-passing iterations are needed; for neighbors three hops away, three iterations are required, and so on. Figure A.1 illustrates how message passing in GNN works.
The message-passing procedure involves two key steps: \textbf{AGGREGATE} and \textbf{UPDATE}. In the \textbf{AGGREGATE} step, the features of neighboring nodes are collected using permutation-invariant functions, such as summation, mean, or maximum. Subsequently, during the \textbf{UPDATE} step, this aggregated information is combined with the current node's state to produce an updated embedding through a neural network layer, often a Multi-Layer Perceptron (MLP). These steps are outlined in Algorithm \ref{alg:message_passing}.

\begin{algorithm}[!t]
\caption{Message Passing in GNNs}
\label{alg:message_passing} 
\begin{algorithmic}[1]
\STATE \textbf{Input:} Graph \( G = (V, E) \), where \( V \) is the set of nodes, and \( E \) is the set of edges. Node features \( \mathbf{h}_v \) for each node \( v \in V \). Maximum iterations \( T \).
\STATE \textbf{Output:} Updated node representations \( \mathbf{h}_v^{(T)} \) for each node \( v \in V \).
\FOR{iteration \( t = 1 \) to \( T \)}
    \FOR{each node \( v \in V \)}
        \STATE \( \mathbf{m}_v^{(t)} = \text{AGGREGATE} \left( \{ \mathbf{h}_u^{(t-1)} : u \in N(v) \} \right) \)
        \STATE \hspace{1cm} \textit{where \( N(v) \) is the set of neighboring nodes of \( v \)}
        \STATE \( \mathbf{h}_v^{(t)} = \text{UPDATE} \left( \mathbf{h}_v^{(t-1)}, \mathbf{m}_v^{(t)} \right) \)
    \ENDFOR
\ENDFOR
\STATE \textbf{Return:} Updated node representations \( \mathbf{h}_v^{(T)} \) for each node \( v \in V \).
\end{algorithmic}
\end{algorithm}

\subsubsection{Graph convolution}
Message passing in the previous section can be implemented in various ways depending on the type of GNN used. Graph convolution, inspired by the convolutional operation in grid-like data (e.g., images) \cite{krizhevsky2012imagenet} is the most popular form of message passing. It involves a structured message passing operation through matrix multiplication between the learnable weight matrix, the node feature matrix, and the adjacency matrix of the input graph \cite{kipf2016semi}. Formally, graph convolution is defined as:

\begin{equation}
H^{(l+1)} = \sigma \left( \tilde{A} H^{(l)} W^{(l)} \right)
\end{equation}

where \( \tilde{A} = D^{-\frac{1}{2}} A D^{-\frac{1}{2}} \) is the normalized adjacency matrix, \( D \) is the degree matrix (a diagonal matrix where each diagonal entry \( D_{ii} \) represents the degree of node \( i \)), \( H^{(l)} \) is the feature matrix at layer \( l \), and \( W^{(l)} \) represents the weight matrix for that layer \cite{kipf2016semi}. This operation captures multi-hop information within each GCN layer. However, deeper GCN layers may encounter the issue of over-smoothing, which occurs when node representations become indistinguishable due to excessive graph Laplacian smoothing during aggregation \cite{li2018deeper,yang2020revisiting}.

\begin{figure}[!t]
    \centering
    \includegraphics[width=0.5\textwidth]{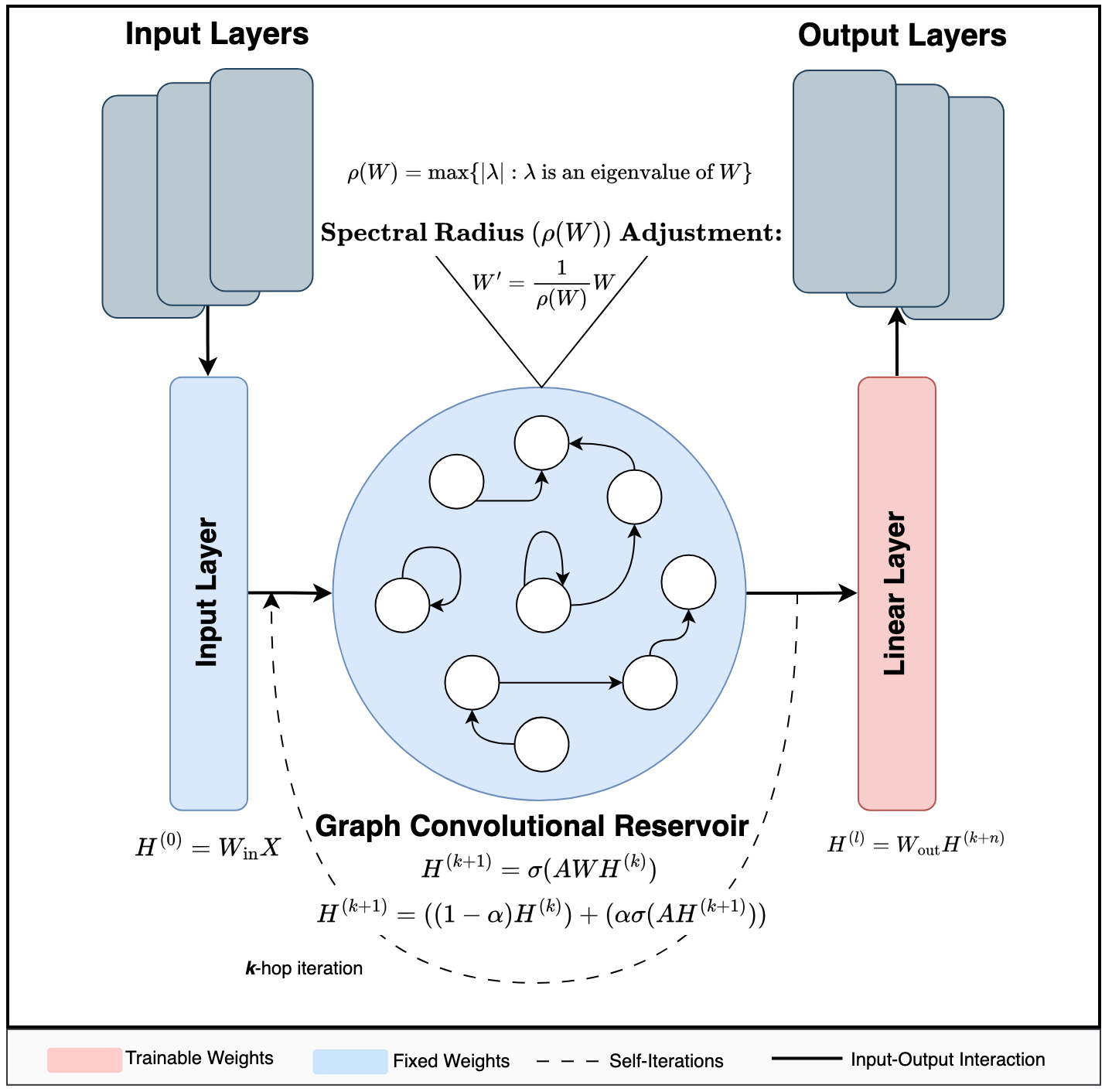}
    \caption{\textit{Proposed RGC-Net architecture, comprising three components: a non-trainable input layer, a graph convolutional reservoir, and a trainable linear output layer.}}
    \label{fig:rgcnet}
\end{figure}

\section{Proposed Reservoir-Based Graph Convolutional Network (RGC-Net)}
The proposed Reservoir-Based Graph Convolutional Network (RGC-Net) leverages the principles of both graph convolution and reservoir computing to iteratively update node embeddings using fixed random weights. Inspired by the architecture of Echo State Networks (ESNs), RGC-Net comprises three primary components: an input layer, a graph convolutional reservoir, and a linear output layer. Each layer plays a unique role in the model architecture, ensuring efficient feature transformation and structured message passing. Figure \ref{fig:rgcnet} illustrates the structure of a single RGC-Net layer, detailed as follows:

\subsection{Input Layer}
The input layer in RGC-Net transforms the initial node features using fixed, random weights, similar to how weight matrices operate in conventional GCN layers. This transformation enables dimensional adjustment, either reducing high-dimensional input features to optimize computation or expanding low-dimensional features for enhanced expressiveness. Formally, the transformation is defined as:
\begin{equation}
H^{(0)} = W_{\text{in}}^{(l)} X,
\label{eq:input_transform}
\end{equation}
where \( H^{(0)} \in \mathbb{R}^{n_h \times n_n} \) represents the transformed node embeddings, \( W_{\text{in}}^{(l)} \in \mathbb{R}^{n_h \times n_f} \) is the fixed input weight matrix at layer \( l \), and \( X \in \mathbb{R}^{n_f \times n_n} \) is the matrix of initial node features. Here, \( n_n \) denotes the number of nodes, \( n_f \) the dimension of initial features, and \( n_h \) the hidden features dimension. This flexible design allows RGC-Net to balance feature expressiveness and computational efficiency by adjusting feature dimensions as needed.
The input layer exists only in the first RGC-Net layer. So, even if there are multiple RGC-Net layers, there will be just one input layer. An illustration of RGC-Net two layers can be found in Figure A.2 in Appendix A.

\subsection{Proposed Reservoir-Based Graph Convolution}
The core of RGC-Net is its \emph{graph convolutional reservoir}, which iteratively refines node embeddings through matrix multiplication with the graph normalized adjacency matrix \( \tilde{A}\in\mathbb{R}^{n_{\text{n}} \times n_n} \) and the fixed, random reservoir weights \( W_{\text{res}}^{(l)}\in\mathbb{R}^{n_{\text{res}} \times n_{\text{res}}} \) (mathematical notations are summarized in Table \ref{tab:notations}). This iterative update enables multi-hop neighborhood information aggregation and is formalized as:
\begin{equation}
H^{(k+1)} = \tilde{A} H^{(k)} W_{\text{res}}^{(l)},
\label{eq:reservoir_update}
\end{equation}
Here, \( H^{(k)}\in\mathbb{R}^{n_n \times n_{\text{res}}} \) is the node embedding at the \( k \)-th iteration. To maintain stability and ensure convergence in the graph convolutional reservoir, the spectral radius of \( W_{\text{res}}^{(l)} \) is adjusted (using Algorithm \ref{alg:spectral_radius_adjustment}) after random initialization. This contraction property stabilizes node state dynamics and promotes effective message passing.
To counteract over-smoothing, a leaky integrator mechanism is integrated into the reservoir, allowing partial retention of initial embeddings. We express this as follows:
\begin{equation}
H^{(k+1)} = \text{ReLU}\left( (1 - \alpha) H^{(k)} + \alpha \tilde{A} H^{(k)} W_{\text{res}}^{(l)} \right),
\label{eq:leaky_integrator_rgc}  
\end{equation}
where \( \alpha \) is the leaky rate controlling the balance between the initial embedding and the aggregated neighborhood information. Lower values of \( \alpha \) preserves more of the target node's initial embedding during node update, while a larger \( \alpha \) preserves less.

Since we are aiming in our work to test \colorbox{myhighlight}{\emph{Hypothesis 1}} regarding the power of non-trainable reservoir weights in graph learning, we also developed a variant of the RGC-Net layer with learnable reservoir weights. This allows us to compare the behaviour of the graph convolutional reservoir with both fixed and learnable weights. The variant with learnable reservoir weights is called TRGC-Net (Trainable Reservoir-based Graph Convolutional Network).

\subsection{Linear Output Layer}
The final output layer in RGC-Net is a simple linear layer with a learnable weight matrix. This matrix transforms the final graph embedding from the reservoir layer obtained after \( n \) iteration \( H^{(k+n)} \) into the desired output dimensions. The transformation at the output layer is defined by:
\begin{equation}
H^{(l)} = W_{\text{out}}^{(l)} H^{(k+n)},
\label{eq:output_layer}
\end{equation}
where \( W_{\text{out}}^{(l)} \in \mathbb{R}^{n_{\text{out}} \times n_{\text{res}}} \) is the output weight matrix that maps node embeddings from the hidden dimension \( n_{\text{res}} \) to the output dimension \( n_{\text{out}} \). This linear layer provides flexibility for RGC-Net to be applied to both classification and generative tasks, as it translates the embeddings learned by the reservoir into task-specific outputs.

In the following sections, we will explore RGC-Net theoretical properties to gain a deeper understanding of its behavior. We will investigate its permutation property to assess its robustness when dealing with the same graph but with different node orders. Furthermore, we will analyze its time and memory complexity to evaluate its theoretical resource requirements.

\subsection{Permutation Properties of RGC-Net}

Permutation invariance and equivariance are essential properties for machine learning models on graph-structured data, especially in graph neural networks (GNNs). Permutation invariance ensures that the model's output remains unchanged under any permutation of the input nodes, while permutation equivariance means that the output reflects the same permutation applied to the input \cite{shawe1993symmetries}. In GNNs, a model is said to be permutation invariant if it yields the same embedding for a graph regardless of the ordering of nodes. Conversely, it is permutation equivariant if the output embedding preserves the structure of any permutation applied to the input graph \cite{hamilton2020graph}.

For RGC-Net, consider a graph permutation defined by \( \tilde{\mathbf{A}}_{\text{perm}} = \mathbf{P} \tilde{\mathbf{A}} \mathbf{P}^{T} \), where \( \mathbf{P} \) is a permutation matrix. We examine the behavior of RGC-Net's graph convolutional reservoir layer under such node permutations. Specifically, the graph convolution operation in RGC-Net will be permutation invariant if \( f(\mathbf{P} \mathbf{x}) = f(\mathbf{x}) \) and permutation equivariant if \( f(\mathbf{P} \mathbf{x}) = \mathbf{P} f(\mathbf{x}) \).
\\

\noindent \textbf{Lemma 1.} \emph{RGC-Net is permutation equivariant. Given a graph permutation defined by \( \tilde{\mathbf{A}}_{\text{perm}} = \mathbf{P} \tilde{\mathbf{A}} \mathbf{P}^{T} \), where \( \mathbf{P} \) is a permutation matrix, the output of RGC-Net's graph convolutional reservoir layer will undergo the same permutation as the input.}

\begin{proof}
To demonstrate the permutation equivariance of RGC-Net, consider the following derivation for the node embeddings at iteration \( k+1 \):

\begin{align*}
    \mathbf{H}_{\text{perm}}^{(k+1)} &= \text{ReLU} \left( (1-\alpha) \cdot \mathbf{P} \mathbf{H}^{(k)} + \alpha \cdot \left( \mathbf{P} \tilde{\mathbf{A}} \mathbf{P}^T \right) \cdot \mathbf{P} \mathbf{H}^{(k)} \right) \\
    &= \text{ReLU} \left( (1-\alpha) \cdot \mathbf{P} \mathbf{H}^{(k)} + \alpha \cdot \mathbf{P} \tilde{\mathbf{A}} \mathbf{H}^{(k)} \mathbf{W} \right) \\
    &= \text{ReLU} \left( \mathbf{P} \left( (1-\alpha) \mathbf{H}^{(k)} + \alpha \tilde{\mathbf{A}} \mathbf{H}^{(k)} \mathbf{W} \right) \right) \\
    &= \mathbf{P} \cdot \text{ReLU} \left( (1-\alpha) \mathbf{H}^{(k)} + \alpha \tilde{\mathbf{A}} \mathbf{H}^{(k)} \mathbf{W} \right) \\
    &= \mathbf{P} \mathbf{H}^{(k+1)}
\end{align*}

This derivation confirms that the RGC-Net graph convolutional reservoir layer exhibits permutation equivariance: when a permutation is applied to the input, the output undergoes the same permutation.
\end{proof}

However, RGC-Net does not demonstrate permutation invariance, as it does not generate identical embeddings for different node orderings, further demonstration can be found in Appendix B. This behavior aligns with traditional GCNs, which are also permutation equivariant~\cite{kipf2016semi}.

\subsection{RGC-Net for Graph Classification}

\textbf{Problem Statement.} Given a set of graphs \( G = \{G_1, G_2, \ldots, G_N\} \) and their corresponding labels \( Y = \{y_1, y_2, \ldots, y_N\} \), where each graph \( G_i \) is defined by an adjacency matrix \( A_i \) and a node feature matrix \( X_i \), i.e., \( G_i = (A_i, X_i) \). The objective is to learn a function \( f : (A_i, X_i) \rightarrow y_i \) that accurately predicts the label \( y_i \) by obtaining a graph representation \( H_{G_i} \) and minimizing a classification loss \( \mathcal{L} \).

\begin{figure*}[!t]
    \centering
    \includegraphics[width=0.84\textwidth]{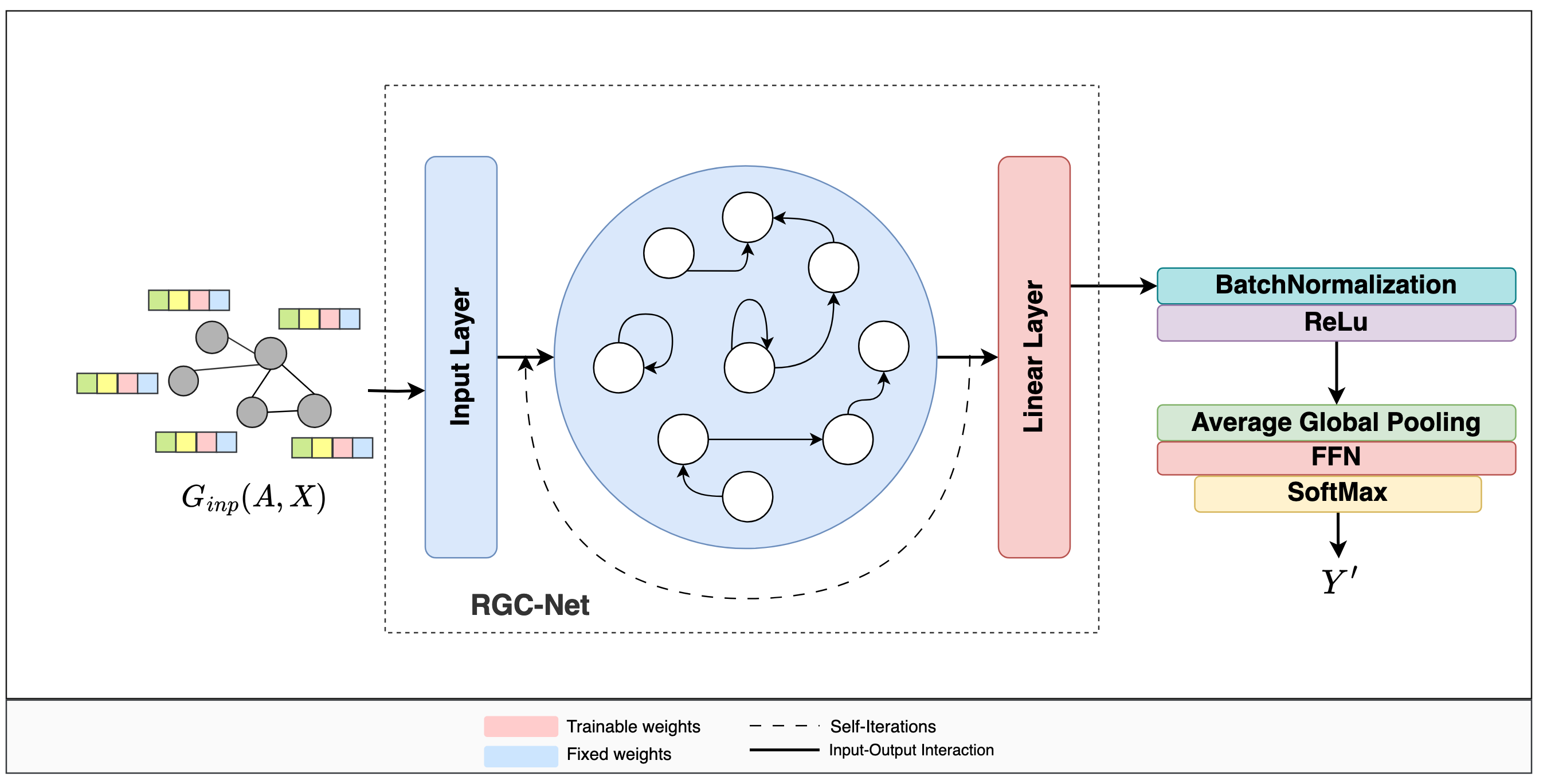}
    \caption{\textit{RGC-Net Architecture for Graph Classification. The input graph \( G_{\text{inp}} \), defined by its adjacency matrix \( A \) and node feature matrix \( X \), is processed through RGC-Net. It begins with an \textbf{Input Layer} that transforms node features using fixed weights. This is followed by the \textbf{Graph Convolutional Reservoir}, where iterative self-iterations capture neighborhood information. The \textbf{Linear Layer} then projects these embeddings to a higher feature space, followed by \textbf{Batch Normalization}, \textbf{ReLU}, \textbf{Average Global Pooling}, and a \textbf{Feed-Forward Network (FFN)}. The \textbf{SoftMax} layer produces the final output label \( Y' \)}}
    \label{fig:classification}
\end{figure*}

\textbf{Graph classification task.} In our research, we develop graph classification models based on the RGC-Net layer and its trainable variant (TRGC-Net) to assign labels to input graphs. The RGC-Net layer generates node embeddings by performing \( k \)-iterations of graph convolution within a reservoir, normalizing the embeddings, and applying a ReLU non-linear activation function. To regularize these transformed node embeddings, we apply batch normalization. The non-linear activation function enhances the model expressive capacity, capturing complex relationships among node features. The resulting node embeddings are then globally pooled using a mean function to produce a comprehensive representation of the entire graph \cite{bruna2014spectral}. This global representation is subsequently passed through a linear layer and a softmax function to yield probability scores for each label, with the predicted label corresponding to the highest probability. The RGC-Net graph classification pipeline is illustrated in Figure \ref{fig:classification}.

The loss function minimized during training, representing the learning objective for our graph classification models, is the negative log-likelihood, formulated as follows:

\begin{equation}
\mathcal{L} = - \log \hat{p}_y
\label{eq:classification_loss}
\end{equation}

where \( \hat{p}_y \) denotes the predicted probability of the correct label \( y \).

\subsection{RGC-Net for Graph Generation}

\textbf{Problem Statement.} Given a sequence of graphs over time \( G = \{G^1, G^2, \ldots, G^T\} \), where each graph is defined as \( G^t = (A^t, X^t) \), the objective is to learn a function \( f : (A^t, X^t) \rightarrow A^{t+1} \) which can predict the adjacency matrix \( A^{t+1} \) for the next time step. By minimizing a predefined loss function \( \mathcal{L} \) over time, the model learns patterns in the structural and feature evolution of the graph, enabling it to generate accurate predictions for future states of the graph.

\begin{figure*}[!t]
    \centering
    \includegraphics[width=0.7\textwidth]{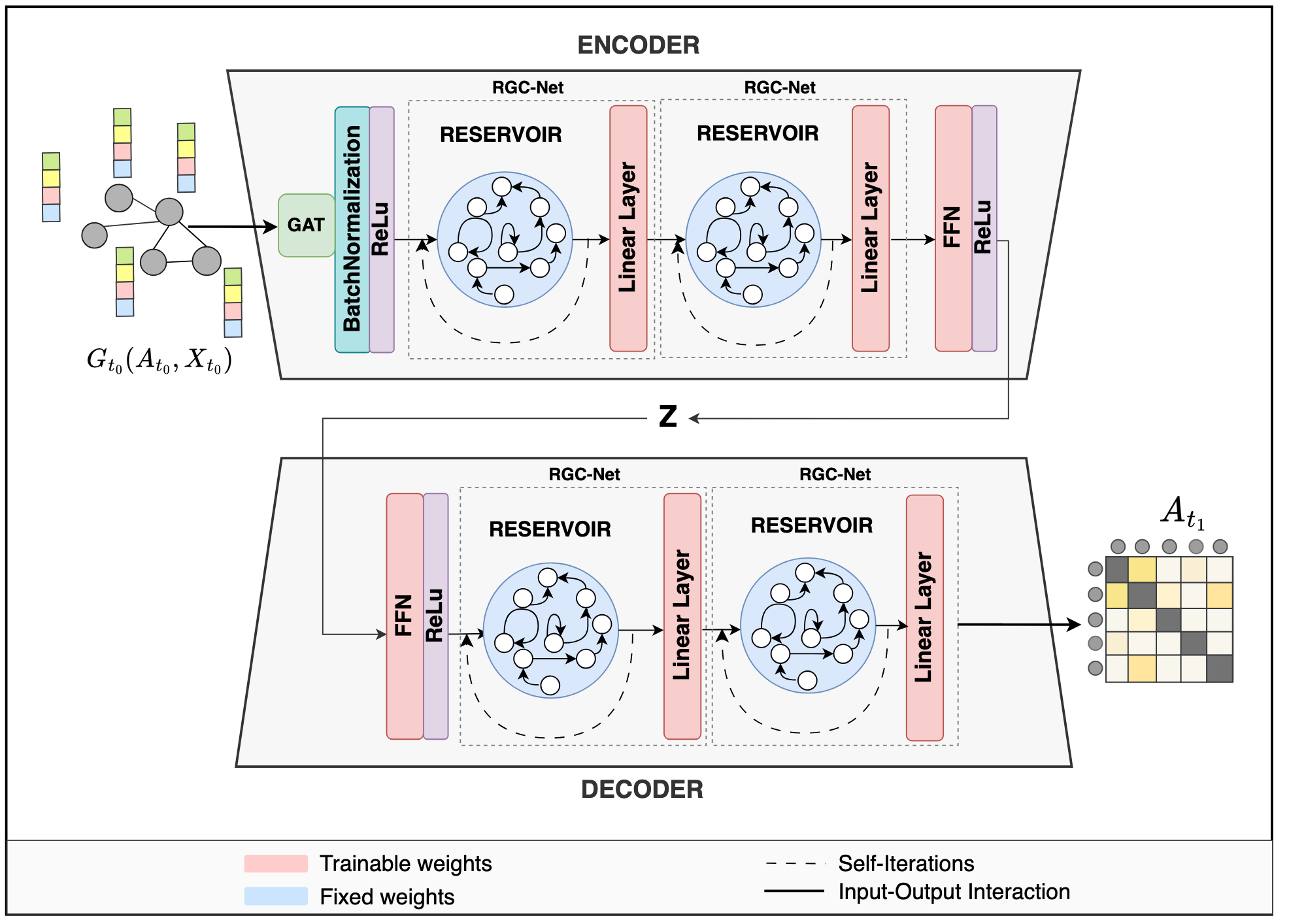}
    \caption{\textit{Architecture of the RGC-Net-Transformer model for temporal graph generation. The model employs RGC-Net layers within encoder and decoder networks to capture dynamic connectivity patterns and predict future graph structures. The encoder processes the input graph \( G_{t_0}(A_{t_0}, X_{t_0}) \) through Graph Attention (GAT), Batch Normalization, and RGC-Net layers, generating a latent representation \( Z \). This is then decoded to reconstruct the predicted adjacency matrix \( A_{t_1} \), utilizing reservoir-based layers to retain features across time steps.}}
    \label{fig:generation}
\end{figure*}

\textbf{Temporal Graph Generation Task.} To develop a flexible reservoir-based convolution framework suitable for various graph learning tasks, we implement our proposed RGC-Net for temporal graph generation, with a focus on predicting brain graph evolution. Specifically, we design a graph transformer network comprising two RGC-Net layers in both the encoder and decoder networks, as shown in Figure \ref{fig:generation}.

In the graph transformer model depicted in Figure \ref{fig:generation}, the RGC-Net input layer is modified to better fit the transformer structure. In the encoder network, the RGC-Net input layer is replaced with a Graph Attention Network (GAT) layer \cite{velivckovic2017graph}, followed by a batch normalization layer. Inspired by the transformer architecture in machine translation \cite{vaswani2017attention}, the GAT layer applies attention mechanisms to emphasize important node features, while batch normalization prevents covariate shifts in the node embeddings after transformation \cite{ioffe2015batch}.

In the decoder network, a simple feed-forward neural (FFN) layer replaces the RGC-Net input layer, transforming the one-dimensional latent representation $Z$ generated by the encoder into a node feature matrix. The RGC-Net layers in the decoder then learn the features of the target graph using node features derived from the latent representation $Z$ and an adjacency matrix initialized as an identity matrix. Finally, the output from the last RGC-Net layer is used to generate the symmetric adjacency matrix of the target predicted graph.

\subsubsection{Loss function}

The learning objective of our generative model, RGC-Net-Transformer, is to minimize the loss function \( \mathcal{L} \), which consists of three components: $L1$ loss (\( \mathcal{L}_{L1} \)), topological loss (\( \mathcal{L}_{top} \)), and Frobenius loss (\( \mathcal{L}_{fro} \)). The influence of each loss term can be adjusted using weights \( \lambda_1 \), \( \lambda_2 \), and \( \lambda_3 \). Thus, the total loss function is defined as follows:

\begin{equation}
\mathcal{L} = \lambda_1 \mathcal{L}_{L1} + \lambda_2 \mathcal{L}_{top} + \lambda_3 \mathcal{L}_{fro}
\label{eq:loss}
\end{equation}

The terms $L1$ loss and topological loss are adapted from recent brain evolution prediction models, such as RBGM \cite{tekin2021recurrent}, 4D-FED-GNN+ \cite{gurler2022federated}, and FedGmTE-Net++ \cite{pistos2023federated}. The $L1$ loss measures the connectivity difference between the actual and generated adjacency matrices $A^{t}_{s}$ and $\hat{A}^{t}_{s}$ of subject $s$ at timepoint $t$ using the mean absolute error (MAE). Since it is based on the mean of the absolute differences between the matrices, $L1$ loss is robust to outliers. The $L1$ loss is mathematically defined as:

\begin{equation}
\mathcal{L}_{L1} = \frac{1}{n} \sum_{i=1}^{n} \left| A^{t}_{s} - \hat{A}^{t}_{s} \right|
\label{eq:l1_loss}
\end{equation}

While the $L1$ loss term ensures that the generated adjacency matrix maintains the connectivity weights of the ground truth adjacency matrix, the topological loss preserves the topological properties of the graph, particularly in terms of node strength. This means that nodes with many or strong connections in the original adjacency matrix will likely retain similar connections in the generated graph. The topological loss term is defined as:

\begin{equation}
\mathcal{L}_{top} = \frac{1}{n} \sum_{i=1}^{n} \left| \sum_{j=1}^{n} A_{ij} - \sum_{j=1}^{n} \hat{A}_{ij} \right|
\label{eq:topological_loss}
\end{equation}
where $A_{ij}$ represent the weight of the edge between node $i$ and node $j$.

In addition, we include Frobenius loss to further encourage the model ability to capture the similarity between the ground truth and generated adjacency matrices in Euclidean space. We define it as:

\begin{equation}
\mathcal{L}_{fro} = \| A^{t}_{s} - \hat{A}^{t}_{s} \|_{F} = \sqrt{ \sum_{i=1}^{m} \sum_{j=1}^{n} \left( A_{ij} - \hat{A}_{ij} \right)^2 }
\label{eq:frobenius_loss}
\end{equation}

\begin{table}[!t]
\centering
\caption{Major Mathematical Notations}
\setlength{\tabcolsep}{3pt}
\renewcommand{\arraystretch}{1}
\resizebox{\columnwidth}{!}{%
\begin{tabular}{|c|c|}
\hline
\textbf{Notation} & \textbf{Definition} \\ \hline \hline
\rule{0pt}{10pt}
$G_s^t = \{A_s^t, X_s^t\}$ & graph of subject $s$ at time point $t$ consists as pairs  \\
& of adjacency matrix $A_s^t$ and node feature matrix $X_s^t$ \\
$A_s^t$ & adjacency matrix of subject $s$ at time point $t$ \\
$\hat{A}_s^t$ & predicted adjacency matrix of subject $s$ at time point $t$ \\
$A_{ij}$ & weight of the edge between node $i$ and node $j$ \\
$X_s^t$ & node features \\
$\mathcal{N}(i)$ & set of neighbouring nodes for node $i$ \\
$n_s$ & number of subjects \\
$n_t$ & number of time points \\
$n_n$ & number of nodes \\
$n_{in}$ & input dimension in the input layer \\
$n_{res}$ & number of neurons in the reservoir  \\
$H^l$ & hidden layer of GCN at layer $l$ \\
$H_l^{(k)}$ & hidden layer of RGC-Net at layer $l$ in the $k$ iteration \\
$\sigma$ & activation function \\
$\alpha$ & reservoir leaky rate \\
$W^{(l)}$ & trainable weight at layer $l$ \\
$W_{in}$ & non-trainable weight of RGC-Net input layer \\
$W_{res}^{(l)}$ & non-trainable weight of reservoir \\
$W_{out}^{(l)}$ & weight of RGC-Net output layer\\
$\mathcal{L}$ & loss function \\
$\mathcal{L}_{l1}$ & L1 loss term of the graph generative models \\
$\mathcal{L}_{top}$ & topological loss term of the graph generative models \\
$\mathcal{L}_{fro}$ & Frobenius loss term of the graph generative models \\
$\lambda_1$ & adjustable weighted coefficient for $\mathcal{L}_{L1}$ \\
$\lambda_2$ & adjustable weighted coefficient for $\mathcal{L}_{top}$ \\
$\lambda_3$ & adjustable weighted coefficient for $\mathcal{L}_{fro}$ \\
$Z$ & latent space\\
\hline
\end{tabular}}
\label{tab:notations}
\end{table}

\section{Experiment Design and Results}

\subsection{Experimental Design}
All models in our experiments are trained inductively using a nested 3-fold cross-validation scheme. In each outer fold, one subset (approximately 33\%) is held out as the test set, while the remaining two subsets (approximately 67\%) are used for training and validation. Within this inner portion, the data are further divided into about 90\% for training and 10\% for validation, which are used for early stopping and hyperparameter tuning. To ensure unbiased results, we conduct cross-validation for RGC-Net models using three separate initializations, each designed to account for any effect of the random initialization of the reservoir's non-trainable weights. Therefore, the performance results for RGC-Net models represent the average of these three runs.

As each model may require a different number of epochs to converge, we use an early stopping mechanism, setting a maximum of 500 epochs for graph classification models and 200 epochs for generative models, with a patience threshold of 5 epochs. This approach stops training when validation loss stabilizes, helping to prevent overfitting. Additionally, we implement a learning rate scheduler, which reduces the learning rate by 50\% after a specified number of epochs. All other hyperparameters are optimized using a grid search, with the grid options detailed in Table \ref{tab2}.

\begin{table}[!t]
    \centering
    \caption{Hyperparameter values for model selection via grid search}
    \setlength{\tabcolsep}{6pt}
    \renewcommand{\arraystretch}{1.1}
    \small
    \begin{tabular}{|c|c|}
        \hline
        \textbf{Hyperparameter} & \textbf{Grid Values} \\
        \hline \hline
        Learning rate (\textit{lr}) & 0.01, 0.005, 0.001 \\
        Scheduler step size & 500, 200, 100 \\
        Reservoir iteration* (\(k\)) & 3, 2, 1 \\
        Reservoir leaky rate* (\(\alpha\)) & 1, 0.9, 0.8, 0.7 \\
        \hline
    \end{tabular}
    \label{tab2}
    \vspace{0.5em}
    \par\small *only for RGC-Net models
\end{table}

\vspace*{-4mm}

\subsection{Graph Classification Experiments}

In this section, we outline our experimental design to evaluate the classification performance of our proposed reservoir-based graph convolution method. We also analyze the impact of various hyperparameters on the model performance.

\subsubsection{Evaluation datasets}

For graph classification, we used three widely employed datasets for binary graph classification: MUTAG \cite{schlichtkrull2018modeling}, PROTEINS \cite{morris2020tudataset, dobson2003distinguishing}, and DD \cite{morris2020tudataset, dobson2003distinguishing}. Each dataset contains homogeneous, static graphs with node features, each has unique classes, nodes, and edge distributions, as shown in Figure A.3 in Appendix A. These datasets exhibit different topologies, providing a means to test the generalizability of the RGC-Net classification model across varied graph structures. The general and topological properties of these datasets are summarized in Appendix C (Table~\ref{Appendix:tab1}).

Each dataset has unique structural properties essential to understanding their topological variations, including \textit{average degree distribution} which indicates the average number of connections (edges) per node, \textit{average clustering coefficient} which reflects the tendency of nodes to form local clusters, \textit{average diameter}, representing the mean of the longest shortest paths between any two nodes, \textit{average density} which measures the connectivity of the graph, and \textit{average modularity} which evaluates how effectively the graph divides into distinct communities or modules. Detailed descriptions of the three graph classification datasets can be found in Appendix C.

\subsubsection{Classification model benchmark}

We evaluate the performance of our proposed RGC-Net model and its trainable variant (TRGC-Net) against two widely used GNN architectures: the Graph Convolutional Network (GCN) \cite{kipf2016semi} and the Graph Attention Network (GAT) \cite{velivckovic2017graph}. To ensure a consistent comparison, we replace the RGC-Net layer with GCN and GAT layers, keeping other architectural elements---such as batch normalization and pooling---unchanged. This setup allows us to directly assess the benefits of introducing a reservoir layer within a standard GNN pipeline.

Our approach is the first to leverage the reservoir computing paradigm specifically to redefine convolution operations within GNNs. Consequently, it is logical and relevant to benchmark RGC-Net against established GNN models like GCN and GAT, as these represent widely adopted frameworks for graph convolutions and attention mechanisms. The goal of our study is to evaluate the impact of a reservoir-based layer in the context of graph convolutions, examining how it enhances performance compared to traditional GNN convolution methods.

\subsubsection{Effect of number of layers}

This experiment evaluates the effect of varying the number of layers on the graph classification performance of RGC-Net compared to other methods. The results, presented in Table \ref{tab4}, demonstrate that RGC-Net consistently achieves high accuracy across different datasets, often outperforming fully trainable models like GAT, GCN, and its trainable variant, TRGC-Net, supporting \colorbox{myhighlight}{\emph{Hypothesis 1}}. On the large DD dataset, RGC-Net scalability becomes evident as it outperforms GAT, which encounters out-of-memory (OOM) errors. This illustrates RGC-Net memory efficiency, as its non-trainable reservoir weights require less memory, a crucial advantage for large-scale graphs. The evaluation also includes reservoir-based and modern graph models such as GraphESN \cite{Gallicchio2010GraphEcho}, GraphSAGE \cite{hamilton2017inductive}, and GIN \cite{xu2018powerful}. Across all datasets, RGC-Net demonstrates superior or comparable performance to these approaches, confirming its strong generalization capability. In particular, RGC-Net achieves notably higher accuracy than GraphESN on the PROTEINS and DD datasets, reflecting the effectiveness of the proposed reservoir design in enhancing representational richness without increasing computational cost. Compared to GraphSAGE and GIN, which rely on trainable neighborhood aggregation and non-linear transformations, RGC-Net attains competitive results with substantially fewer trainable parameters.

Additionally, RGC-Net exhibits a more gradual decline in accuracy with increased layers, indicating better control over the over-smoothing effect. This supports \colorbox{myhighlight}{\emph{Hypothesis 2}}, showing that RGC-Net can better control over-smoothing by preserving the initial node embedding using the leaky rate $(\alpha)$. Overall, these findings reinforce RGC-Net's strengths in both efficiency and discriminative capacity, establishing it as a robust and scalable alternative to conventional and reservoir-based GNNs alike.

\begin{table*}[!t]
    \centering
    \caption{The accuracies of graph classification models with different numbers of layers across three datasets. The results showing the highest accuracy are highlighted in \textbf{\textcolor{blue}{blue}}. Models marked with \textcolor{red}{OOM} encountered an out-of-memory error.}
    \setlength{\tabcolsep}{6pt}
    \renewcommand{\arraystretch}{1.3}
    \small
    \begin{tabular}{|c|c|c|c|c|c|c|c|c|}
        \hline
        \textbf{Dataset} & \textbf{n-layers} & \textbf{GAT} & \textbf{GCN} & \textbf{GraphSAGE} & \textbf{GIN} & \textbf{GraphESN} & \textbf{RGC-Net} & \textbf{TRGC-Net} \\ \hline \hline

        \multirow{5}{*}{\rotatebox{90}{\textbf{MUTAG}}}
        & 1 & 0.7446$\pm$0.04 & 0.7500$\pm$0.01 & 0.7446$\pm$0.04 & 0.8300$\pm$0.06 & 0.6433$\pm$0.0071 & \textbf{\textcolor{blue}{0.8726$\pm$0.01}} & 0.8671$\pm$0.05 \\
        & 2 & 0.7715$\pm$0.07 & 0.8355$\pm$0.05 & 0.7024$\pm$0.08 & 0.7710$\pm$0.04 & 0.6479$\pm$0.0064 & \textbf{\textcolor{blue}{0.8600$\pm$0.01}} & 0.8462$\pm$0.09 \\
        & 3 & 0.7661$\pm$0.03 & 0.8297$\pm$0.01 & 0.5856$\pm$0.19 & 0.6608$\pm$0.24 & 0.6579$\pm$0.0075 & 0.8458$\pm$0.02 & \textbf{\textcolor{blue}{0.8461$\pm$0.06}} \\
        & 4 & 0.7502$\pm$0.04 & 0.7977$\pm$0.03 & 0.7607$\pm$0.07 & \textbf{\textcolor{blue}{0.8455$\pm$0.08}} & 0.6579$\pm$0.0075 & 0.8317$\pm$0.04 & \textbf{\textcolor{blue}{0.8405$\pm$0.02}} \\
        & 5 & 0.7606$\pm$0.03 & 0.7873$\pm$0.01 & 0.7714$\pm$0.04 & 0.8244$\pm$0.05 & 0.6579$\pm$0.0075 & 0.8032$\pm$0.02 & \textbf{\textcolor{blue}{0.8297$\pm$0.06}} \\ \hline

        \multirow{5}{*}{\rotatebox{90}{\textbf{PROTEINS}}}
        & 1 & 0.5229$\pm$0.11 & 0.6020$\pm$0.03 & 0.5642$\pm$0.09 & 0.5957$\pm$0.03 & 0.5919$\pm$0.0029 & \textbf{\textcolor{blue}{0.6038$\pm$0.01}} & 0.5337$\pm$0.08 \\
        & 2 & 0.5004$\pm$0.09 & 0.6056$\pm$0.01 & 0.4681$\pm$0.10 & 0.4591$\pm$0.10 & 0.5937$\pm$0.0014 & \textbf{\textcolor{blue}{0.6101$\pm$0.00}} & 0.5597$\pm$0.05 \\
        & 3 & 0.4906$\pm$0.10 & 0.6002$\pm$0.02 & 0.4034$\pm$0.02 & 0.6029$\pm$0.02 & 0.5912$\pm$0.0048 & \textbf{\textcolor{blue}{0.6050$\pm$0.01}} & 0.5660$\pm$0.07 \\
        & 4 & \textbf{\textcolor{blue}{0.6083$\pm$0.02}} & 0.6029$\pm$0.01 & 0.4124$\pm$0.04 & 0.5615$\pm$0.09 & 0.5949$\pm$0.0006 & 0.5981$\pm$0.00 & 0.5768$\pm$0.04 \\
        & 5 & 0.5948$\pm$0.02 & 0.5759$\pm$0.02 & 0.4483$\pm$0.10 & \textbf{\textcolor{blue}{0.5993$\pm$0.03}} & 0.5949$\pm$0.0011 & 0.5978$\pm$0.00 & 0.5894$\pm$0.03 \\ \hline

        \multirow{5}{*}{\rotatebox{90}{\textbf{DD}}}
        & 1 & \textcolor{red}{OOM} & \textbf{\textcolor{blue}{0.6052$\pm$0.05}} & 0.6188$\pm$0.04 & 0.3939$\pm$0.01 & 0.5836$\pm$0.0014 & 0.5605$\pm$0.01 & 0.4694$\pm$0.08 \\
        & 2 & \textcolor{red}{OOM} & 0.5602$\pm$0.04 & 0.4313$\pm$0.07 & 0.4889$\pm$0.09 & \textbf{\textcolor{blue}{0.5799$\pm$0.0050}} & 0.5684$\pm$0.02 & 0.5000$\pm$0.08 \\
        & 3 & \textcolor{red}{OOM} & 0.5136$\pm$0.09 & 0.5840$\pm$0.06 & 0.4694$\pm$0.02 & 0.5824$\pm$0.0033 & \textbf{\textcolor{blue}{0.5846$\pm$0.02}} & 0.4584$\pm$0.09 \\
        & 4 & \textcolor{red}{OOM} & 0.5254$\pm$0.04 & 0.4168$\pm$0.03 & 0.3837$\pm$0.01 & \textbf{\textcolor{blue}{0.5848$\pm$0.0018}} & 0.5637$\pm$0.04 & 0.4881$\pm$0.10 \\
        & 5 & \textcolor{red}{OOM} & 0.5067$\pm$0.04 & 0.5272$\pm$0.10 & 0.4118$\pm$0.03 & 0.5862$\pm$0.0002 & \textbf{\textcolor{blue}{0.5874$\pm$0.00}} & 0.5866$\pm$0.03 \\ \hline
    \end{tabular}
    \label{tab4}
\end{table*}

\subsubsection{Effect of iterations and leaky rate}

To evaluate the influence of the iteration count (\( k \)) and leaky rate (\( \alpha \)) on RGC-Net classification performance, we conduct experiments using various configurations of \( k \) and \(\alpha\). The datasets used were MUTAG, a generic graph dataset with inherent node features, and EMCI-AD, a connectomic graph dataset without inherent node features. For MUTAG, we classify the graph labels, while for EMCI-AD, we classify the time point labels of brain graphs from two different time. In addition, we conducted an extended sensitivity analysis (with accompanying plot (Figure A.4) provided in Appendix A) to further examine the robustness of RGC-Net under a wider range of $k$ and $\alpha$ values.

As shown in Table \ref{tab5}, RGC-Net demonstrates strong performance across different settings, generally surpassing its trainable variant (TRGC-Net), further supporting \colorbox{myhighlight}{\emph{Hypothesis 1}}. This suggests that RGC-Net fixed reservoir dynamics effectively capture sufficient information for robust graph representations without the additional overhead of trainable weights. Additionally, the results reveal an interesting pattern in the performance of the RGC-Net model on the MUTAG and EMCI-AD datasets, depending on the number of iterations $(k)$ and the leaky rate $(\alpha)$. On the MUTAG dataset, RGC-Net achieves its highest average accuracy of \(0.8611 \pm 0.01\) with fewer iterations (\(k = 1\)). This suggests that the MUTAG dataset, which comprises smaller and simpler chemical compound graphs, benefits from more localized neighborhood aggregation. The high accuracy at \(k = 1\) indicates that a single iteration of message passing is sufficient to capture the necessary structural patterns, while additional iterations may introduce redundant or noisy information, potentially diluting the specificity of node embeddings. Conversely, on the EMCI-AD dataset, the highest average accuracy is achieved at \(k = 4\) iterations, with an accuracy of \(0.5120 \pm 0.02\) for RGC-Net. This result implies that the EMCI-AD dataset, which contains brain connectivity graphs with more complex structures, requires deeper message passing to capture meaningful relationships. The increased number of iterations allows the model to aggregate information from a wider neighborhood, which appears crucial for representing the intricate, non-local dependencies within brain connectivity networks.

The results also show that TRGC-Net (trainable reservoir) often does not outperform RGC-Net (fixed reservoir), particularly on the MUTAG dataset where the highest average accuracy for TRGC-Net is \(0.8539 \pm 0.02\) at $k=2$, slightly below RGC-Net's best. This suggests that the fixed, non-trainable reservoir weights in RGC-Net provide sufficient representation power while also contributing to faster convergence and resource efficiency, which supports \colorbox{myhighlight}{\emph{Hypothesis 3}}. However, TRGC-Net performs competitively on EMCI-AD, where trainability may offer a slight advantage in learning complex structures. This could be because the added flexibility allows TRGC-Net to adapt its weights specifically to the unique characteristics of the EMCI-AD dataset, though it comes with an increased computational cost.

\begin{table*}[!t]
\centering
\caption{Accuracies of RGC-Net and its variant (TRGC-Net) on graph classification with MUTAG and EMCI-AD datasets for various iterations (\(k\)) and leaky rates (\(\alpha\)). The highest accuracy in each iteration is bolded in \textcolor{blue}{\textbf{blue}} and the best average accuracy across all iterations is bolded in \textcolor{darkgreen}{\textbf{green}}.}
   \setlength{\tabcolsep}{6pt}
    \renewcommand{\arraystretch}{1.2}
    \small
\begin{tabular}{|c|c|cc|cc|}
\hline
\multirow{2}{*}{\begin{tabular}[c]{@{}c@{}}n-iteration\\ (\(k\))\end{tabular}} & \multirow{2}{*}{\begin{tabular}[c]{@{}c@{}}leaky\\ rate (\(\alpha\))\end{tabular}} & \multicolumn{2}{c|}{\textbf{MUTAG}} & \multicolumn{2}{c|}{\textbf{EMCI-AD}} \\ \cline{3-6}
 &  & \textbf{RGC-Net*} & \textbf{TRGC-Net} & \textbf{RGC-Net*} & \textbf{TRGC-Net} \\ \hline \hline
1 & 1 & 0.8442$\pm$0.01 & 0.8087$\pm$0.06 & 0.4476$\pm$0.03 & 0.4476$\pm$0.01 \\
1 & 0.9 & 0.8584$\pm$0.01 & \textcolor{blue}{\textbf{0.8564$\pm$0.05}} & 0.4504$\pm$0.03 & \textcolor{blue}{\textbf{0.4847$\pm$0.08}} \\
1 & 0.8 & \textcolor{blue}{\textbf{0.8726$\pm$0.01}} & 0.8408$\pm$0.07 & \textcolor{blue}{\textbf{0.4573$\pm$0.02}} & 0.4475$\pm$0.05 \\
1 & 0.7 & 0.8691$\pm$0.01 & 0.8245$\pm$0.03 & 0.4302$\pm$0.01 & 0.4625$\pm$0.02 \\ \rowcolor{gray!20} \hline
\textbf{Avg.} &  & \textcolor{darkgreen}{\textbf{0.8611$\pm$0.01}} & 0.8207$\pm$0.02 & 0.4464$\pm$0.01 & 0.4606$\pm$0.02 \\ \hline
2 & 1 & 0.8335$\pm$0.01 & 0.8247$\pm$0.03 & \textcolor{blue}{\textbf{0.5149$\pm$0.03}} & \textcolor{blue}{\textbf{0.4998$\pm$0.04}} \\
2 & 0.9 & 0.8441$\pm$0.00 & \textcolor{blue}{\textbf{0.8671$\pm$0.05}} & 0.4902$\pm$0.02 & 0.4921$\pm$0.04 \\
2 & 0.8 & 0.8564$\pm$0.01 & 0.8618$\pm$0.03 & 0.4927$\pm$0.03 & 0.4702$\pm$0.00 \\
2 & 0.7 & \textcolor{blue}{\textbf{0.8618$\pm$0.01}} & 0.8618$\pm$0.07 & 0.4925$\pm$0.01 & 0.4614$\pm$0.12 \\ \rowcolor{gray!20}\hline
\textbf{Avg.} &  & 0.8490$\pm$0.01 & \textcolor{darkgreen}{\textbf{0.8539$\pm$0.02}} & 0.4976$\pm$0.01 & 0.4809$\pm$0.02 \\ \hline
3 & 1 & 0.8228$\pm$0.02 & 0.7926$\pm$0.02 & 0.4777$\pm$0.02 & 0.4476$\pm$0.05 \\
3 & 0.9 & 0.8282$\pm$0.03 & 0.8192$\pm$0.03 & 0.4724$\pm$0.03 & 0.4855$\pm$0.04 \\
3 & 0.8 & \textcolor{blue}{\textbf{0.8476$\pm$0.02}} & 0.8193$\pm$0.05 & 0.4762$\pm$0.02 & 0.4623$\pm$0.05 \\
3 & 0.7 & 0.8316$\pm$0.03 & \textcolor{blue}{\textbf{0.8618$\pm$0.06}} & \textcolor{blue}{\textbf{0.4823$\pm$0.02}} & \textcolor{blue}{\textbf{0.5673$\pm$0.02}} \\ \rowcolor{gray!20}\hline
\textbf{Avg.} &  & 0.8326$\pm$0.01 & 0.8232$\pm$0.03 & 0.4763$\pm$0.01 & \textcolor{darkgreen}{\textbf{0.4981$\pm$0.05}} \\ \hline
4 & 1 & 0.8157$\pm$0.02 & 0.7767$\pm$0.03 & 0.5150$\pm$0.05 & 0.4177$\pm$0.04 \\
4 & 0.9 & 0.7981$\pm$0.02 & 0.8085$\pm$0.05 & 0.4976$\pm$0.02 & 0.4545$\pm$0.08 \\
4 & 0.8 & 0.8050$\pm$0.04 & 0.8245$\pm$0.05 & 0.4977$\pm$0.03 & 0.4401$\pm$0.02 \\
4 & 0.7 & \textcolor{blue}{\textbf{0.8423$\pm$0.03}} & \textcolor{blue}{\textbf{0.8298$\pm$0.05}} & \textcolor{blue}{\textbf{0.5375$\pm$0.01}} & \textcolor{blue}{\textbf{0.4699$\pm$0.03}} \\ \rowcolor{gray!20} \hline
\textbf{Avg.} &  & 0.8153$\pm$0.02 & 0.8099$\pm$0.02 & \textcolor{darkgreen}{\textbf{0.5120$\pm$0.02}} & 0.4456$\pm$0.02 \\ \hline
5 & 1 & 0.7802$\pm$0.03 & 0.7873$\pm$0.01 & 0.4924$\pm$0.01 & 0.4625$\pm$0.02 \\
5 & 0.9 & 0.7996$\pm$0.01 & \textcolor{blue}{\textbf{0.8190$\pm$0.04}} & \textcolor{blue}{\textbf{0.5073$\pm$0.05}} & 0.4700$\pm$0.03 \\
5 & 0.8 & 0.8156$\pm$0.01 & 0.7876$\pm$0.05 & 0.4973$\pm$0.04 & \textcolor{blue}{\textbf{0.4848$\pm$0.05}} \\
5 & 0.7 & \textcolor{blue}{\textbf{0.8297$\pm$0.01}} & 0.7600$\pm$0.13 & 0.4802$\pm$0.02 & 0.4771$\pm$0.05 \\ \rowcolor{gray!20} \hline
\textbf{Avg.} &  & 0.8063$\pm$0.02 & 0.7885$\pm$0.02 & 0.4943$\pm$0.01 & 0.4736$\pm$0.01 \\ \hline
\end{tabular}
\label{tab5}
\end{table*}

\subsubsection{Resource consumption}
The proposed RGC-Net exhibits substantially reduced computational overhead, particularly on larger graphs. As shown in Table~\ref{tab:computational_efficiency}, RGC-Net achieves the lowest training time, fastest per-epoch execution, and fewest required epochs across all models on both the PROTEINS and DD datasets, reducing training time by up to 90\% compared to conventional GNNs (e.g., 1.83~s vs.\ 11.84--25.69~s on PROTEINS). This efficiency stems from its non-trainable reservoir, which eliminates backpropagation through the graph convolutional layers, drastically cutting computational overhead. Despite its simplicity, RGC-Net maintains memory and GPU usage comparable to or lower than standard GNNs (e.g., 1.78~GB memory on PROTEINS, matching GCN), confirming its lightweight nature. In contrast, the trainable variant (TRGC-Net) incurs higher costs due to gradient computation, underscoring the advantage of fixed reservoir weights. Overall, RGC-Net offers a compelling trade-off: minimal resource consumption without sacrificing representational capacity, making it especially suitable for large-scale or resource-constrained graph learning scenarios.

\begin{table*}[!t]
\centering
\caption{Comparison of Training Time, Time per Epoch, Number of Epochs, Average Memory Usage, and Average GPU Usage across different models and datasets.}
\setlength{\tabcolsep}{4pt}
\renewcommand{\arraystretch}{1.2}

\begin{tabular}{|c|c|c|c|c|c|c|}
\hline
\textbf{Dataset} & \textbf{Models} & \textbf{Training} & \textbf{Time per} & \textbf{\#Epoch} & \textbf{Avg. Memory} & \textbf{Avg. GPU} \\
                 &                 & \textbf{Time (s)}  & \textbf{Epoch (s)} &                  & \textbf{Usage (GB)}  & \textbf{Usage (GB)} \\ \hline \hline

\multirow{6}{*}{\rotatebox{90}{\textbf{MUTAG}}}
& GAT & $5.11 \pm 3.72$ & $0.030 \pm 0.028$ & $169.27 \pm 175.5$ & $1.6251 \pm 0.0060$ & $0.01678 \pm 0.00007$ \\
& GCN & $5.62 \pm 2.26$ & $0.021 \pm 0.008$ & $273.80 \pm 127.6$ & \textbf{\textcolor{blue}{$1.6391 \pm 0.0022$}} & \textbf{\textcolor{blue}{$0.01690 \pm 0.00007$}} \\
& GraphSAGE & $\mathbf{\textcolor{blue}{2.85 \pm 1.43}}$ & $0.022 \pm 0.011$ & \textbf{\textcolor{blue}{$129.90 \pm 103.5$}} & $1.8790 \pm 0.0255$ & $0.02060 \pm 0.00008$ \\
& GIN & $5.14 \pm 1.13$ & $0.025 \pm 0.005$ & $203.33 \pm 41.7$ & $1.8792 \pm 0.0248$ & $0.02062 \pm 0.00010$ \\
& RGC-NET & $6.65 \pm 2.97$ & $0.020 \pm 0.009$ & $326.91 \pm 151.3$ & $1.6489 \pm 0.0002$ & $\mathbf{\textcolor{blue}{0.01684 \pm 0.00005}}$ \\
& TRGC-NET & $10.00 \pm 7.23$ & $0.039 \pm 0.028$ & $256.80 \pm 158.4$ & $1.7350 \pm 0.0029$ & $\mathbf{\textcolor{blue}{0.01684 \pm 0.00005}}$ \\ \hline

\multirow{6}{*}{\rotatebox{90}{\textbf{PROTEINS}}}
& GAT & $20.96 \pm 20.68$ & $0.226 \pm 0.254$ & $92.73 \pm 135.5$ & $1.8078 \pm 0.0219$ & $0.01744 \pm 0.00011$ \\
& GCN & $11.84 \pm 12.87$ & $0.108 \pm 0.118$ & $109.20 \pm 124.3$ & $1.7837 \pm 0.0000$ & $0.01752 \pm 0.00008$ \\
& GraphSAGE & $25.20 \pm 8.19$ & $0.126 \pm 0.041$ & $200.53 \pm 88.4$ & $1.8932 \pm 0.0001$ & $0.02056 \pm 0.00004$ \\
& GIN & $25.69 \pm 10.45$ & $0.118 \pm 0.048$ & $217.93 \pm 142.7$ & $1.8933 \pm 0.0000$ & $0.02058 \pm 0.00007$ \\
& RGC-NET & $\mathbf{\textcolor{blue}{1.83 \pm 0.28}}$ & $\mathbf{\textcolor{blue}{0.011 \pm 0.002}}$ & $\mathbf{\textcolor{blue}{16.09 \pm 2.58}}$ & $1.7837 \pm 0.0000$ & $\mathbf{\textcolor{blue}{0.01754 \pm 0.00005}}$ \\
& TRGC-NET & $10.54 \pm 9.33$ & $0.206 \pm 0.180$ & $54.20 \pm 78.6$ & $1.7847 \pm 0.0008$ & $0.01762 \pm 0.00011$ \\ \hline

\multirow{5}{*}{\rotatebox{90}{\textbf{DD}}}
& GCN & $82.16 \pm 43.55$ & $0.224 \pm 0.113$ & $366.53 \pm 174.5$ & $1.7964 \pm 0.0002$ & $0.02706 \pm 0.00011$ \\
& GraphSAGE & $24.47 \pm 22.54$ & $0.196 \pm 0.178$ & $125.20 \pm 126.6$ & $1.8933 \pm 0.0000$ & $0.02252 \pm 0.0010$ \\
& GIN & $28.91 \pm 22.37$ & $0.187 \pm 0.146$ & $154.40 \pm 135.1$ & $1.8933 \pm 0.0000$ & $0.02286 \pm 0.0013$ \\
& RGC-NET & $\mathbf{\textcolor{blue}{7.73 \pm 4.76}}$ & $\mathbf{\textcolor{blue}{0.050 \pm 0.031}}$ & $\mathbf{\textcolor{blue}{31.96 \pm 21.5}}$ & $1.7969 \pm 0.0001$ & $\mathbf{\textcolor{blue}{0.02668 \pm 0.00007}}$ \\
& TRGC-NET & $38.92 \pm 35.33$ & $0.441 \pm 0.397$ & $88.26 \pm 88.6$ & $1.7997 \pm 0.0026$ & $0.02708 \pm 0.00035$ \\ \hline
\end{tabular}
\label{tab:computational_efficiency}
\end{table*}

\subsection{Graph Generation Experiments}
In this section, we evaluate the generative performance of our proposed RGC-Net model in predicting brain graph evolution, specifically targeting connectivity dynamics in longitudinal datasets. We also analyze RGC-Net computational efficiency compared to established generative models and its trainable variant, TRGC-Net.

\begin{figure}[!t]
    \centering
    \includegraphics[width=0.5\textwidth]{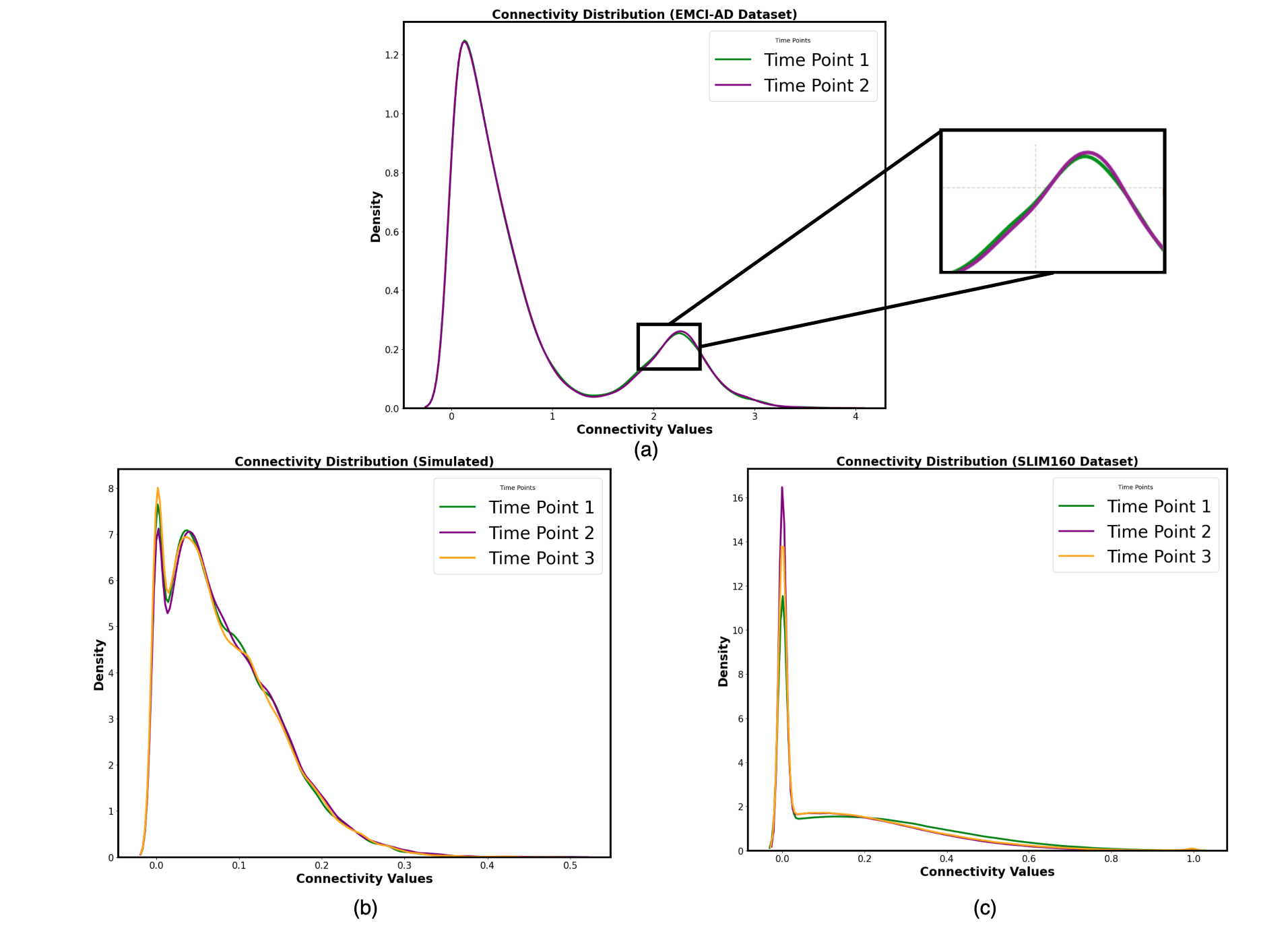}
    \caption{\textit{Connectivity distributions of connectomic datasets for brain graph evolution prediction across different time points. Figure (a) shows the EMCI-AD dataset with nearly the same connectivity distribution across time points. Figures (b) and (c) show connectivity distribution for simulated and SLIM160 datasets. The figures show that although the connectivity distribution between time points is quite similar, each time point slightly differs in the maximum density.}}
    \label{fig:connectivity_distribution}
\end{figure}

\subsubsection{Evaluation datasets}
The datasets used to evaluate the generative capabilities of RGC-Net include two real connectomic datasets (EMCI-AD \cite{mueller2005alzheimer} and SLIM160 \cite{liu2017longitudinal}) and a simulated dataset generated following the longitudinal connectomic data simulation approach in \cite{demirbilek2023predicting}. These datasets consist of temporal graphs representing brain connectivity over time, derived from MRI imaging data, and are predominantly sparse in their connectivity structure. Further details of the connectomic datasets can be found in Appendix C, as well as their graph properties in (Table~\ref{Appendix:tab2}).

Figure \ref{fig:connectivity_distribution} illustrates the connectivity distributions across different time points for each dataset, revealing relatively stable patterns over time. In the EMCI-AD dataset (Figure \ref{fig:connectivity_distribution}(a)), connectivity distributions between time points are nearly identical, with only minor variations, particularly in maximum density. This observation aligns with neuroscience findings, which indicate that structural brain changes in patients with early Alzheimer's and Mild Cognitive Impairment are typically observable on an annual basis rather than within a six-month interval \cite{coupe2019lifespan,song2013dynamics}. Similarly, the simulated dataset (Figure \ref{fig:connectivity_distribution}(b)) and the SLIM160 dataset (Figure \ref{fig:connectivity_distribution}(c)) exhibit consistent distribution patterns across time points, with slight density fluctuations at each time point. These patterns support the neuroscientific view that structural brain connectivity changes minimally in healthy adults over time \cite{resnick2000one,raz2005regional}.

Overall, these consistent yet slightly varying connectivity distributions across datasets suggest that temporal patterns are preserved, providing a suitable basis for evaluating RGC-Net ability to generalize temporal connectivity changes effectively.

\begin{table*}[!t]
\centering
\caption{Evaluation results for brain graph evolution prediction across three longitudinal connectomic datasets. The best results for each dataset are highlighted in \textbf{\textcolor{blue}{blue}}. Models marked with \textcolor{red}{OOM} encountered an out-of-memory error.}
\setlength{\tabcolsep}{4pt}
\renewcommand{\arraystretch}{1.2}

\begin{tabular}{|c|c|c|c|c|c|c|c|}
\hline
\textbf{Dataset} & \textbf{Models} & \textbf{$\downarrow$MAE} & \textbf{$\downarrow$Frobenius} & \textbf{$\downarrow$MAE} & \textbf{$\downarrow$MAE} & \textbf{$\downarrow$MAE} & \textbf{$\downarrow$MAE} \\
\textbf{}        &                 &                          & \textbf{Distance}              & \textbf{Node}            & \textbf{Clustering}      & \textbf{Betweenness}     & \textbf{Eigenvector} \\
\textbf{}        &                 &                          &                                & \textbf{Strength}        & \textbf{Coef.}           & \textbf{C.}              & \textbf{C.} \\ \hline \hline
\multirow{6}{*}{\rotatebox{90}{\textbf{EMCI-AD}}}
                & Identity Function        & 0.0860$\pm$0.01 & 3.9414$\pm$0.54  & 1.3765$\pm$0.24  & 0.0101$\pm$0.00 & 0.1051$\pm$0.00 & 0.0048$\pm$0.00 \\
                & RBGM \cite{tekin2021recurrent}  & 0.2516$\pm$0.08 & 11.3652$\pm$3.77 & 7.2024$\pm$2.78  & 0.0740$\pm$0.00 & 0.0915$\pm$0.00 & \textbf{\textcolor{blue}{0.0093$\pm$0.00}} \\
                & EvoGraphNet \cite{nebli2020deep} & 0.4135$\pm$0.02 & 23.3185$\pm$0.61 & 11.1199$\pm$0.30 & 0.2106$\pm$0.02 & 0.0916$\pm$0.01 & 0.0221$\pm$0.00 \\
                & GCN-Transformer          & 0.6847$\pm$0.56 & 39.2852$\pm$34.8 & 20.4643$\pm$10.7 & \textbf{\textcolor{blue}{0.0436$\pm$0.02}} & 0.1026$\pm$0.01 & 0.0136$\pm$0.00 \\
                & RGC-Net-Transformer      & \textbf{\textcolor{blue}{0.1860$\pm$0.02}} & \textbf{\textcolor{blue}{9.0352$\pm$1.13}} & \textbf{\textcolor{blue}{2.5498$\pm$0.30}} & 0.0468$\pm$0.01 & 0.0954$\pm$0.00 & 0.0106$\pm$0.00 \\
                & TRGC-Net-Transformer     & 0.2115$\pm$0.04 & 10.6004$\pm$1.59 & 3.3258$\pm$0.85  & 0.0594$\pm$0.01 & \textbf{\textcolor{blue}{0.0913$\pm$0.00}} & 0.0120$\pm$0.00 \\ \hline
\multirow{6}{*}{\rotatebox{90}{\textbf{Simulated}}}
                & Identity Function        & 0.0422$\pm$0.00 & 1.9740$\pm$0.09 & 0.7076$\pm$0.04 & 0.0581$\pm$0.00 & 0.2722$\pm$0.01 & 0.0274$\pm$0.00 \\
                & RBGM \cite{tekin2021recurrent}  & 0.0374$\pm$0.00 & 1.7078$\pm$0.01 & 0.6868$\pm$0.03 & 0.1748$\pm$0.01 & 0.2398$\pm$0.01 & 0.0240$\pm$0.00 \\
                & EvoGraphNet \cite{nebli2020deep} & 0.0441$\pm$0.01 & 2.0467$\pm$0.26 & 0.7974$\pm$0.15 & 0.0749$\pm$0.01 & 0.0749$\pm$0.01 & 0.0296$\pm$0.00 \\
                & GCN-Transformer          & 0.0353$\pm$0.00 & 1.6243$\pm$0.09 & 0.6046$\pm$0.03 & \textbf{\textcolor{blue}{0.0731$\pm$0.01}} & \textbf{\textcolor{blue}{0.2311$\pm$0.01}} & 0.0242$\pm$0.00 \\
                & RGC-Net-Transformer      & \textbf{\textcolor{blue}{0.0305$\pm$0.00}} & \textbf{\textcolor{blue}{1.4081$\pm$0.02}} & \textbf{\textcolor{blue}{0.5308$\pm$0.01}} & 0.0884$\pm$0.01 & 0.2404$\pm$0.00 & \textbf{\textcolor{blue}{0.0208$\pm$0.00}} \\
                & TRGC-Net-Transformer     & 0.0316$\pm$0.00 & 1.4488$\pm$0.05 & 0.5374$\pm$0.04 & 0.0882$\pm$0.01 & 0.2432$\pm$0.01 & 0.0210$\pm$0.00 \\ \hline
\multirow{6}{*}{\rotatebox{90}{\textbf{SLIM160}}}
                & Identity Function        & 0.1450$\pm$0.00 & 31.2550$\pm$0.69 & 10.3607$\pm$1.59 & 0.0562$\pm$0.01 & 0.0286$\pm$0.00 & 0.0273$\pm$0.00 \\
                & RBGM \cite{tekin2021recurrent}  & \textcolor{red}{OOM} & \textcolor{red}{OOM} & \textcolor{red}{OOM} & \textcolor{red}{OOM} & \textcolor{red}{OOM} & \textcolor{red}{OOM} \\
                & EvoGraphNet \cite{nebli2020deep} & \textcolor{red}{OOM} & \textcolor{red}{OOM} & \textcolor{red}{OOM} & \textcolor{red}{OOM} & \textcolor{red}{OOM} & \textcolor{red}{OOM} \\
                & GCN-Transformer          & 0.1480$\pm$0.01 & 31.6629$\pm$2.14 & 10.2580$\pm$1.06 & \textbf{\textcolor{blue}{0.0673$\pm$0.01}} & 0.0308$\pm$0.00 & 0.0268$\pm$0.00 \\
                & RGC-Net-Transformer      & 0.1439$\pm$0.00 & 27.7943$\pm$0.80 & 8.3279$\pm$0.64  & 0.4460$\pm$0.05 & \textbf{\textcolor{blue}{0.0245$\pm$0.00}} & 0.0233$\pm$0.00 \\
                & TRGC-Net-Transformer     & \textbf{\textcolor{blue}{0.1425$\pm$0.00}} & \textbf{\textcolor{blue}{27.7208$\pm$0.72}} & \textbf{\textcolor{blue}{8.3297$\pm$0.64}} & 0.4150$\pm$0.04 & 0.0316$\pm$0.00 & \textbf{\textcolor{blue}{0.0232$\pm$0.00}} \\ \hline
\end{tabular}
\label{tab:Evaluation results_evolution}
\end{table*}

\begin{table*}[!t]
\centering
\caption{Comparison of Training Time, Time per Epoch, Number of Epochs, Average Memory Usage, and Average GPU Usage across different models and datasets.}
\setlength{\tabcolsep}{4pt}
\renewcommand{\arraystretch}{1.2}

\begin{tabular}{|c|c|c|c|c|c|c|}
\hline
\textbf{Dataset} & \textbf{Models} & \textbf{Training} & \textbf{Time per} & \textbf{\#Epoch} & \textbf{Avg. Memory} & \textbf{Avg. GPU} \\
                 &                 & \textbf{Time (s)}  & \textbf{Epoch (s)} &                  & \textbf{Usage (GB)}  & \textbf{Usage (GB)} \\ \hline \hline
\multirow{5}{*}{\rotatebox{90}{\textbf{EMCI-AD}}} & RBGM \cite{tekin2021recurrent}  & 176.42$\pm$21.69 & 3.92$\pm$0.69 & 45$\pm$5.73  & 3.31$\pm$0.00 & 0.17$\pm$0.00 \\
                                   & EvoGraphNet \cite{nebli2020deep}  & 260.29$\pm$124.42 & 1.30$\pm$0.62 & 200$\pm$0.00 & 3.26$\pm$0.00 & 0.03$\pm$0.00 \\
                                   & GCN-Transformer   & 8.85$\pm$1.75 & \textbf{\textcolor{blue}{0.29$\pm$0.08}} & 31$\pm$6.48 & \textbf{\textcolor{blue}{3.29$\pm$0.00}} & \textbf{\textcolor{blue}{0.02$\pm$0.00}} \\
                                   & RGC-Net-Transformer & \textbf{\textcolor{blue}{8.76$\pm$0.44}} & 0.30$\pm$0.02 & \textbf{\textcolor{blue}{29$\pm$1.70}} & 4.31$\pm$0.27 & \textbf{\textcolor{blue}{0.02$\pm$0.00}} \\
                                   & TRGC-Net-Transformer & 15.6$\pm$10.0 & 0.33$\pm$0.29 & 47$\pm$28.5 & 4.02$\pm$0.22 & \textbf{\textcolor{blue}{0.02$\pm$0.00}} \\ \hline
\multirow{5}{*}{\rotatebox{90}{\textbf{Simulated}}} & RBGM \cite{tekin2021recurrent} & 409.53$\pm$35.9 & 17.81$\pm$2.23 & 23$\pm$2.05 & 3.27$\pm$0.01 & 0.12$\pm$0.00 \\
                                     & EvoGraphNet \cite{nebli2020deep} & 1410.50$\pm$539.39 & 7.05$\pm$2.70 & 200$\pm$0.00 & 3.24$\pm$0.01 & 0.05$\pm$0.00 \\
                                     & GCN-Transformer   & 39.42$\pm$8.06 & 0.86$\pm$0.25 & 46$\pm$9.42 & \textbf{\textcolor{blue}{3.27$\pm$0.00}} & \textbf{\textcolor{blue}{0.02$\pm$0.00}} \\
                                     & RGC-net-Transformer & \textbf{\textcolor{blue}{30.28$\pm$7.60}} & \textbf{\textcolor{blue}{0.82$\pm$0.26}} & \textbf{\textcolor{blue}{37$\pm$7.54}} & 3.72$\pm$0.00 & \textbf{\textcolor{blue}{0.02$\pm$0.00}} \\
                                     & TRGC-Net-Transformer & 53.02$\pm$2.40 & 1.66$\pm$0.10 & 32$\pm$1.25 & 4.32$\pm$0.00 & \textbf{\textcolor{blue}{0.02$\pm$0.00}} \\ \hline
\multirow{5}{*}{\rotatebox{90}{\textbf{SLIM160}}}   & RBGM \cite{tekin2021recurrent} & \textcolor{red}{OOM} & \textcolor{red}{OOM} & \textcolor{red}{OOM} & \textcolor{red}{OOM} & \textcolor{red}{OOM} \\
                                     & EvoGraphNet \cite{nebli2020deep} & \textcolor{red}{OOM} & \textcolor{red}{OOM} & \textcolor{red}{OOM} & \textcolor{red}{OOM} & \textcolor{red}{OOM} \\
                                     & GCN-Transformer   & 46.0$\pm$5.36 & \textbf{\textcolor{blue}{0.94$\pm$0.15}} & 49$\pm$5.73 & \textbf{\textcolor{blue}{3.76$\pm$0.32}} & \textbf{\textcolor{blue}{0.06$\pm$0.00}} \\
                                     & RGC-Net-Transformer & \textbf{\textcolor{blue}{12.8$\pm$1.72}} & 1.07$\pm$0.20 & \textbf{\textcolor{blue}{12$\pm$1.63}} & 3.81$\pm$0.10 & \textbf{\textcolor{blue}{0.06$\pm$0.00}} \\
                                     & TRGC-Net-Transformer & 104.1$\pm$3.14 & 5.48$\pm$0.40 & 19$\pm$1.25 & 4.17$\pm$0.10 & \textbf{\textcolor{blue}{0.06$\pm$0.00}} \\ \hline
\end{tabular}
\label{tab:comparison_training_time}
\end{table*}

\subsubsection{Evaluation measures}
To evaluate the generative performance of RGC-Net for brain graph evolution, we assess similarity between generated and ground-truth brain graphs based on connectivity properties:

\paragraph{Mean Absolute Error (MAE)} measures the connectivity discrepancy (edge weights) between nodes in generated and ground truth adjacency matrices, defined as:

\begin{equation}
\text{MAE} = \frac{1}{n} \sum_{i=1}^{n} \left| A^{t}_{s} - \hat{A}^{t}_{s} \right|
\end{equation}

\paragraph{Frobenius distance (FD)} quantifies the Euclidean distance between adjacency matrices, treating each matrix as a high-dimensional vector:

\begin{equation}
\text{FD} = \left\| A^{t}_{s} - \hat{A}^{t}_{s} \right\|_F = \sqrt{ \sum_{i=1}^{m} \sum_{j=1}^{n} \left( A_{ij} - \hat{A}_{ij} \right)^2 }
\end{equation}

\paragraph{MAE Node Strength} evaluates preservation of node importance, calculated as the average discrepancy in node strength (sum of edge weights) between generated and ground truth graphs:

\begin{equation}
\text{MAE Node Strength} = \frac{1}{n} \sum_{i=1}^{n} \left| \hat{s}_{i} - s_{i} \right|
\end{equation}
where $s_{i}= \sum_{i=1}^{n} A_{ij}$ depicts the node strength; higher node strength means the node has stronger connections.

\paragraph{MAE Clustering Coefficient} assesses the retention of local clustering characteristics (i.e., local neighborhood density) by comparing the clustering coefficients $C_i$ and $\hat{C}_i$ of the generated and ground truth graphs:
\begin{equation}
\text{MAE Clustering Coefficient} = \frac{1}{n} \sum_{i=1}^{n} \left| C_i - \hat{C}_i \right|
\end{equation}

\paragraph{MAE Betweenness Centrality (BC)}
Another way to quantify a node's importance is by calculating its betweenness centrality \cite{newman2010networks}. Betweenness centrality measures how a node acts as a connector between other nodes, influencing the flow of information across the entire graph. Mathematically, betweenness centrality is described as:
\begin{equation}
C_B(i) = \sum_{s \neq i \neq t} \frac{\sigma_{st}(i)}{\sigma_{st}}
\label{eq:betweenness_centrality}
\end{equation}

where \( \sigma_{st}(i) \) represents the number of shortest paths from node \( s \) to node \( t \) that pass through the node \( i \) and \( \sigma_{st} \) is the total number of shortest paths from \( s \) to \( t \).
To assess how well the generated graph preserves the betweenness centrality of each node from the ground truth graph, we use the MAE of betweenness centrality defined as:

\begin{equation}
\text{MAE BC} = \frac{1}{n} \sum_{i=1}^{n} \left| C_B(i) - \hat{C}_B(i) \right|
\label{eq:mae_betweenness}
\end{equation}

\paragraph{MAE Eigenvector Centrality} reflects a node's influence by considering both direct connections (its degree) and indirect connections (the degrees of its neighbouring nodes) \cite{newman2010networks}. Nodes with high eigenvector centrality are connected to other well-connected nodes, indicating their essential roles within the network. Eigenvector centrality is calculated as:

\begin{equation}
C_E(i) = \frac{1}{\lambda} \sum_{j \in N(i)} A_{ij} C_E(j)
\label{eq:eigenvector_centrality}
\end{equation}

where \( \lambda \) is the largest eigenvalue of the adjacency matrix \( A \) and \( N(i) \) is a set of neighbours of node \( i \). We calculate the MAE of eigenvector centrality between the ground truth and generated graphs to compare the influence of nodes in both graphs defined as:

\begin{equation}
\text{MAE Eigenvector Centrality} = \frac{1}{n} \sum_{i=1}^{n} \left| C_E(i) - \hat{C}_E(i) \right|
\label{eq:mae_eigenvector_centrality}
\end{equation}

\subsubsection{Generative model benchmarks}
To benchmark the generative performance of our proposed RGC-Net-Transformer model, we evaluate it against three methods used for brain graph evolution prediction: the Identity Mapper baseline and two state-of-the-art (SOTA) models, RBGM \cite{tekin2021recurrent} and EvoGraphNet \cite{nebli2020deep}. In addition to these comparisons, we introduce two variants of the RGC-Net-Transformer model for a more comprehensive evaluation. The first variant, GCN-Transformer, replaces the RGC-Net layer with a traditional GCN layer. The second variant, TRGC-Net-Transformer, modifies the RGC-Net layer by making the weights of both the reservoir input layer and the reservoir itself trainable.

\subsubsection{Generative performance comparison}
In this experiment, we evaluate the generative performance of the proposed RGC-Net Transformers model against its variants and other methods for the brain graph evolution prediction task. The results presented in Table \ref{tab:Evaluation results_evolution} demonstrate that the proposed transformer-based models (GCN-Transformers, RGC-Net-Transformers, TRGC-Net-Transformers) outperformed state-of-the-art models (RBGM and EvoGraphNet) across all datasets. Additionally, the transformer-based models demonstrated better scalability on larger temporal graph data (SLIM160) than RBGM and EvoGraphNet, which encountered out-of-memory errors when working with SLIM160. The scalability issues with RBGM and EvoGraphNet are likely due to their edge-based convolution, which requires high memory consumption.
In experiments with the Simulated and SLIM160 datasets, all brain graph evolution prediction models outperformed the identity function baseline. However, no generative models outperformed the identity function for the EMCI-AD dataset, likely due to the highly similar connectivity distribution between time points, as shown in Figure \ref{fig:connectivity_distribution}, making it difficult for the models to learn distinct graph representations for each time point. The RGC-Net-Transformer and TRGC-Net-Transformer exhibit consistent yet distinct performance patterns across datasets. On the Simulated dataset, RGC-Net achieves slightly better MAE, Frobenius distance, MAE node strength, and eigenvector centrality than TRGC-Net, indicating that the fixed reservoir sufficiently captures the necessary graph dynamics without overfitting. On the EMCI-AD dataset, RGC-Net also performs competitively, while TRGC-Net gains minor improvements in betweenness centrality (Table~\ref{tab:Evaluation results_evolution}), reflecting its ability to adapt to more complex or irregular connectivity patterns. For the SLIM160 dataset, both RGC-Net and TRGC-Net remain scalable and outperform GCN-Transformer, with RGC-Net achieving the lowest Frobenius Distance and MAE Node Strength, demonstrating its robustness and efficiency on large temporal graphs. Overall, these results illustrate the trade-off between fixed and trainable reservoirs: while RGC-Net-Transformer (fixed reservoirs) excels in efficiency, scalability, and robustness, particularly on smaller or less complex graphs, TRGC-Net-Transformer (trainable reservoirs) provides marginal improvements on datasets with complex or highly irregular connectivity patterns (e.g., EMCI-AD), but at higher computational cost and slower training.

\subsubsection{Resource consumption comparisons}
In addition to evaluating the generative performance of our proposed model, we monitor its resource consumption during training and compare it to its variants and state-of-the-art models like RBGM and EvoGraphNet.
According to Table \ref{tab:comparison_training_time}, transformer-based models consume less GPU than state-of-the-art models. The transformer models with GCN, RGC-Net, and TRGC-Net layers show similar average GPU usage during training. Furthermore, although the data indicates that RGC-Net Transformer used more memory than GCN-Transformer, RGC-Net-Transformer has faster training time as it requires fewer epochs to converge. However, the training time per epoch for the RGC-Net-Transformer is slightly slower than the GCN-Transformer, which is consistent with the complexity analysis, discussing the higher time and memory costs of the RGC-Net layer compared to the GCN layer.
The faster training time of the RGC-Net model is likely due to the fewer parameters that need to be optimized during training. Moreover, since the reservoir mapping of input features to a high-dimensional space is unaffected by the input features, it introduces a regularization effect that further helps the RGC-Net model to converge more quickly. These results partially support \colorbox{myhighlight}{\emph{Hypothesis 3}}. While the RGC-Net model shows faster convergence with shorter training times, the non-trainable weights in the RGC-Net reservoir do not demonstrate significant efficiency improvements over conventional GCN in terms of resource consumption.

\section{Conclusion and Future Work}
In this work, we introduced \emph{reservoir-based graph convolutions} in GNNs (RGC-Net), the first graph convolution method leveraging the reservoir computing paradigm. By utilizing the recursive, dynamic properties and leaky integrator within reservoir computing, we developed RGC-Net for both graph classification and temporal graph generation, specifically for predicting brain graph evolution.

Our evaluations demonstrate that RGC-Net outperforms established methods like GCN and GAT in graph classification and excels in brain graph evolution prediction, even surpassing recent state-of-the-art (SOTA) approaches. The leaky integrator proves effective in mitigating over-smoothing and preserving essential node features during updates. Furthermore, RGC-Net exhibits faster convergence in temporal graph generation tasks, though resource efficiency improvements are not significant.

While RGC-Net showed promise, it is not permutation-invariant, which may affect its performance on graphs with varying node orders or isomorphic structures. Additionally, the model requires tuning of two additional hyperparameters---leaky rate and iteration count---which increases complexity relative to traditional GCNs. Although RGC-Net achieves faster convergence, further optimization is necessary to improve efficiency for deployment on resource-constrained devices.

Future directions include extending RGC-Net to a wider pool of graph learning tasks such as node classification and edge prediction, broadening the applicability of \emph{reservoir-based graph convolutions}. Additionally, exploring structured reservoir topologies and alternative connectivity patterns could enhance model performance. Investigating efficient methods to tune or perturb the non-trainable reservoir weights, such as direct feedback alignment, may further improve RGC-Net effectiveness. Finally, experimenting with other reservoir types, such as Liquid State Machines and Backpropagation-Decorrelation, could offer new insights and computational advantages.

\bibliographystyle{IEEEtran}
\bibliography{biblio}

\section*{Biography}

\vspace{-9pt}
\begin{IEEEbiography}[{\includegraphics[width=1in,height=1.25in,clip,keepaspectratio]{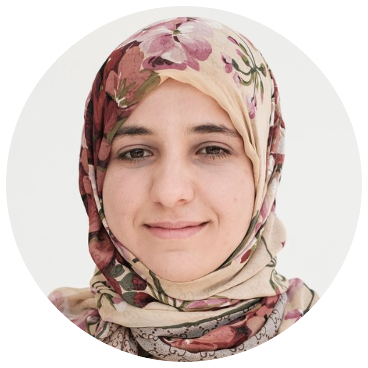}}]{Mayssa Soussia}
Ph.D. student at the National Engineering School of Sousse (ENISo), University of Sousse, Tunisia. She is affiliated with both LATIS (Laboratory of Advanced Technology and Intelligent Systems) and BASIRA (Brain And SIgnal Research and Analysis) laboratories. In 2019--2020, she was a research scholar at IDEA Lab, Department of Radiology and Biomedical Research Imaging Center (BRIC), University of North Carolina at Chapel Hill, Chapel Hill, NC, USA. She earned her Master's degree from the National Engineering School of Tunis (ENIT), University of Tunis El Manar, in 2017, and her Engineering Diploma from the Higher National School of Engineers of Tunis (ENSIT), University of Tunis, in 2015. She has won the MICCAI 2024 Student Travel Award for the African continent and previously won the Best Paper Award at the Connectomics in Neuroimaging Workshop, MICCAI 2017. Since 2019, she has served as a Program Committee (PC) member for the IEEE ISBI conference and several MICCAI workshops, including PRIME and EPIMI.
\end{IEEEbiography}

\begin{IEEEbiography}[{\includegraphics[width=1in,height=1.25in,clip,keepaspectratio]{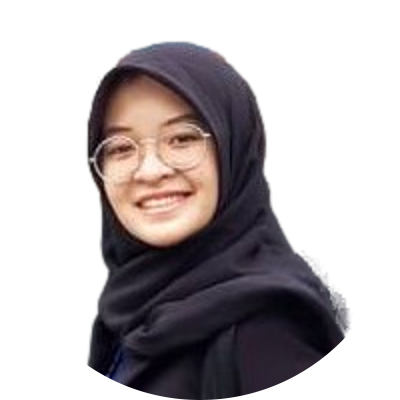}}]{Gita Ayu Salsabila}
MSc student in Computing at Imperial College London, specializing in advanced machine learning and artificial intelligence. Prior to her research work, she served as a Business Intelligence professional at Shopee, Jakarta, Indonesia. She also co-founded BACARA, an innovative sign language translation app that received recognition as one of the top 15 Indonesian capstone projects in the Bangkit program. Gita also gained industry experience as a QA Engineer at Fulldive, California where she developed an advanced URL identifier for the Fulldive Browser. Her expertise spans machine learning, artificial intelligence, data engineering, and software development, with a strong focus on leveraging innovative technologies to solve complex challenges.
\end{IEEEbiography}

\begin{IEEEbiography}[{\includegraphics[width=1in,height=1.25in,clip,keepaspectratio]{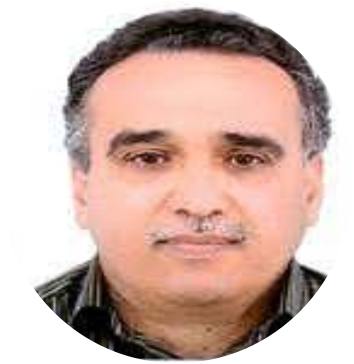}}]{Mohamed Ali Mahjoub}
He received the Bsc and MSc degrees in computer science from the National School of Computer Science of Tunis, Tunisia and National School of Engineers of Tunis, in 1990 and 1993, respectively, and the PhD and HDR degrees in electrical engineering and signal processing from the National School of Engineers of Tunis, Tunisia, in 1999 and 2013, respectively. From 1990 to 1999, he was a computer science engineer with the National School of Engineers of Tunis. In September 1999, he joined the Preparatory Institute of Engineering of Monastir as an assistant professor in computer science. In December 2013, he joined the Electronics Engineering Department, National School of Engineers of Sousse (ENISo) as a professor. He is a founding member of the Laboratory of Advanced Technology and Intelligent Systems (LATIS), ENISo. His research interests include machine learning, probabilistic graphical models, computer vision, and pattern recognition.
\end{IEEEbiography}

\begin{IEEEbiography}[{\includegraphics[width=1in,height=1.25in,clip,keepaspectratio]{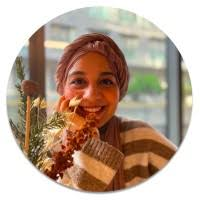}}]{Islem Rekik}
is the Director of the Brain And SIgnal Research and Analysis (BASIRA) laboratory (\url{http://basira-lab.com/}) and an Associate Professor at Imperial College London (Innovation Hub I-X). She is the awardee of two prestigious international research fellowships. In 2019, she was awarded the 3-year prestigious TUBITAK 2232 for Outstanding Experienced Researchers Fellowship and in 2020 she became a Marie Sklodowska-Curie fellow under the European Horizons 2020 program. Together with BASIRA members, she conducted more than 100 cutting-edge research projects cross-pollinating AI and healthcare---with a sharp focus on brain imaging and network neuroscience. She is also a co/chair/organizer of more than 25 international first-class conferences/workshops/competitions (e.g., Affordable AI 2021--22, Predictive AI 2018--2024, Machine Learning in Medical Imaging 2021--24, WILL competition 2021--23). She is a member of the organizing committee of MICCAI 2023 (Vancouver), 2024 (Marrakesh) and South-Korea (2025). She will serve as the General Co-Chair of MICCAI 2026 in Abu Dhabi. In addition to her 160+ high-impact publications, she is a strong advocate of equity, inclusiveness and diversity in AI and research. She is the former president of the Women in MICCAI (WiM), and the co-founder of the international RISE Network to Reinforce Inclusiveness \& diverSity and Empower minority researchers in Low-Middle Income Countries (LMIC).
\end{IEEEbiography}

\newpage

\appendices
\onecolumn
\section{Supplementary figures}
\renewcommand{\thefigure}{A.1}
\begin{figure}[H]
    \centering
    \includegraphics[width=0.6\linewidth]{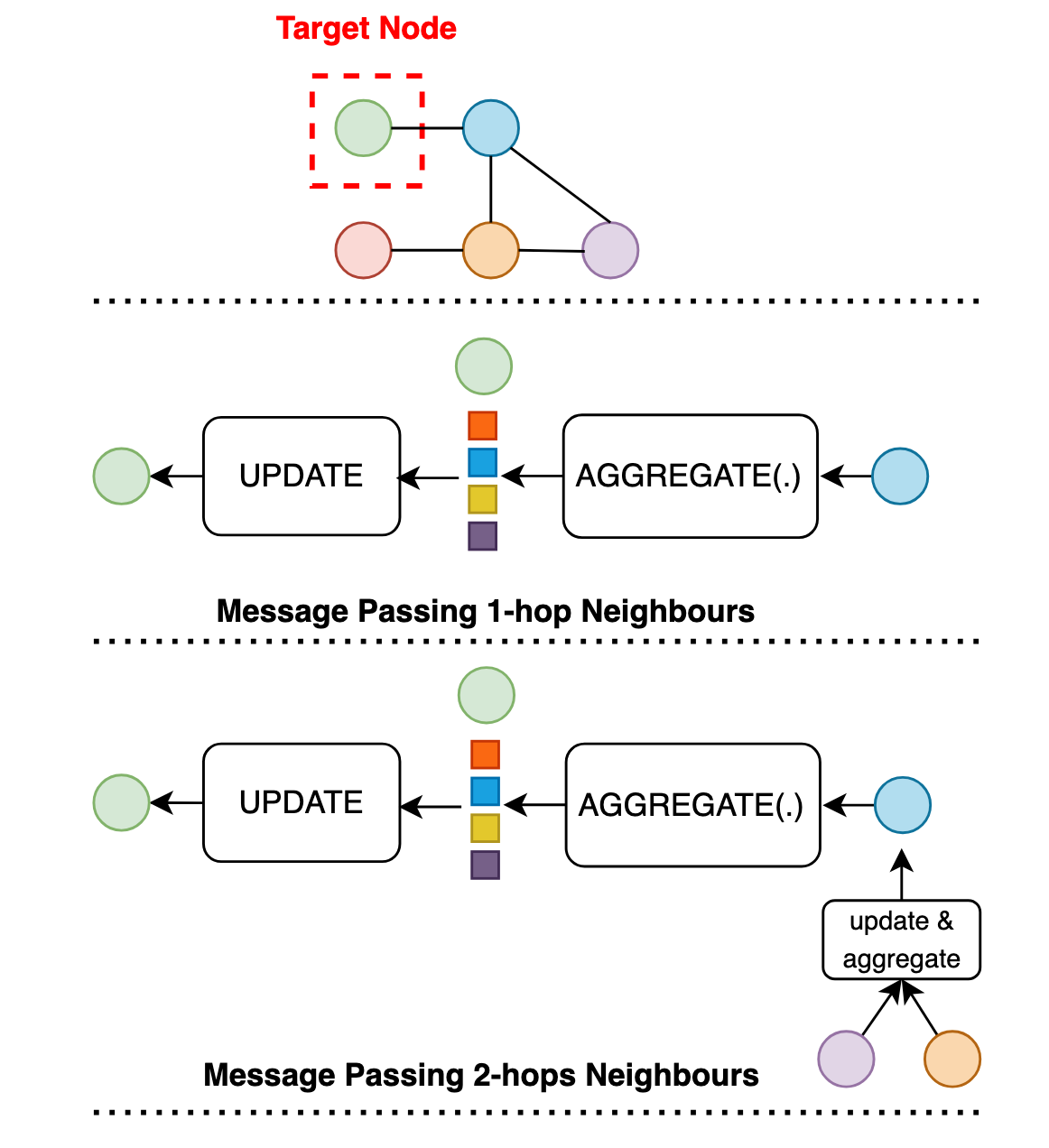}
    \caption{\textit{Message passing operation to update target node embedding}}
    \label{fig:enter-label}
\end{figure}

\renewcommand{\thefigure}{A.2}
\begin{figure}[H]
    \centering
    \includegraphics[width=0.78\textwidth]{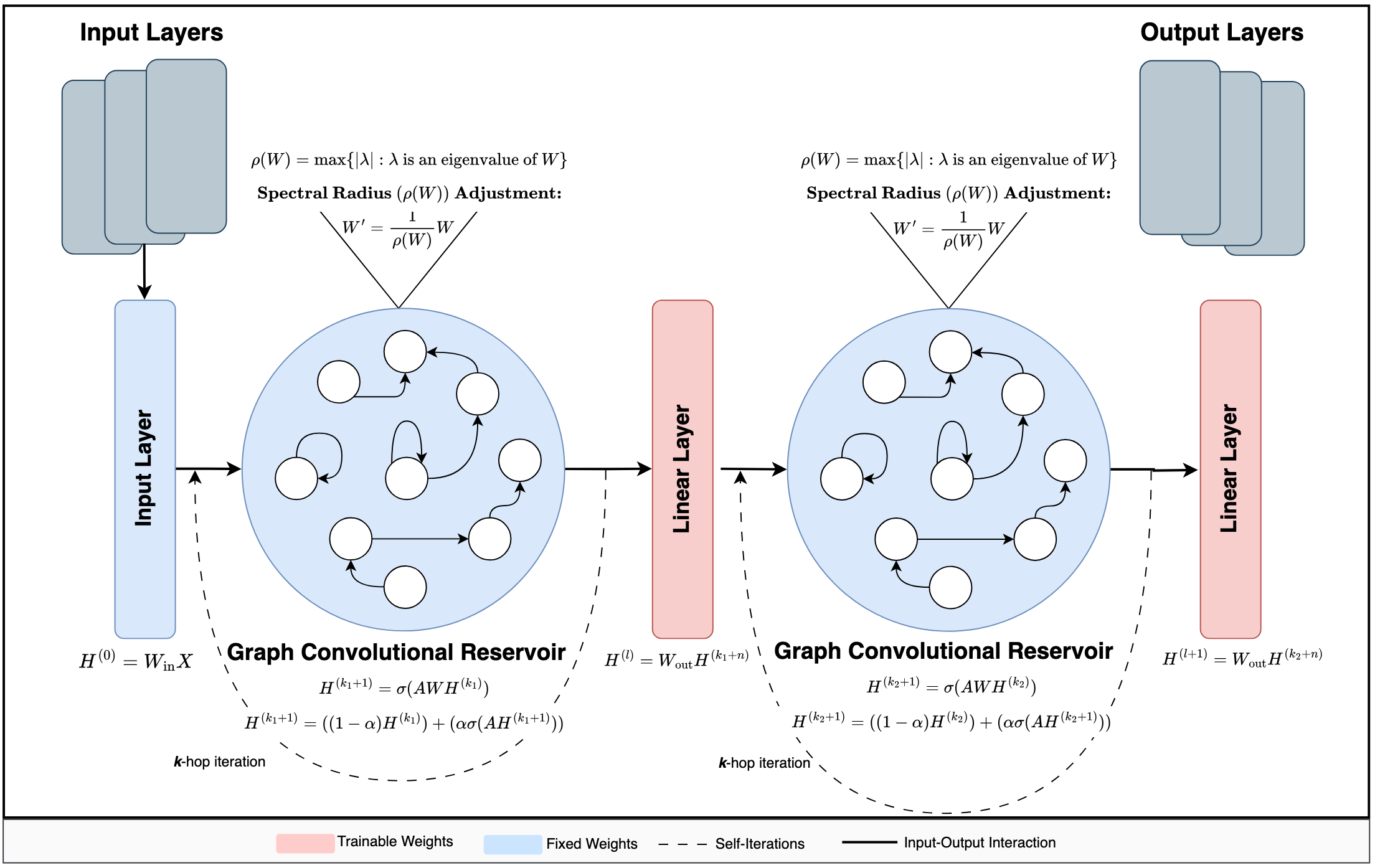}
    \caption{\textit{Illustration of a 2-layer RGC-Net architecture}}
    \label{fig:RGCNet_2L}
\end{figure}

\renewcommand{\thefigure}{A.3}
\begin{figure}[H]
    \centering
    \includegraphics[width=1\linewidth]{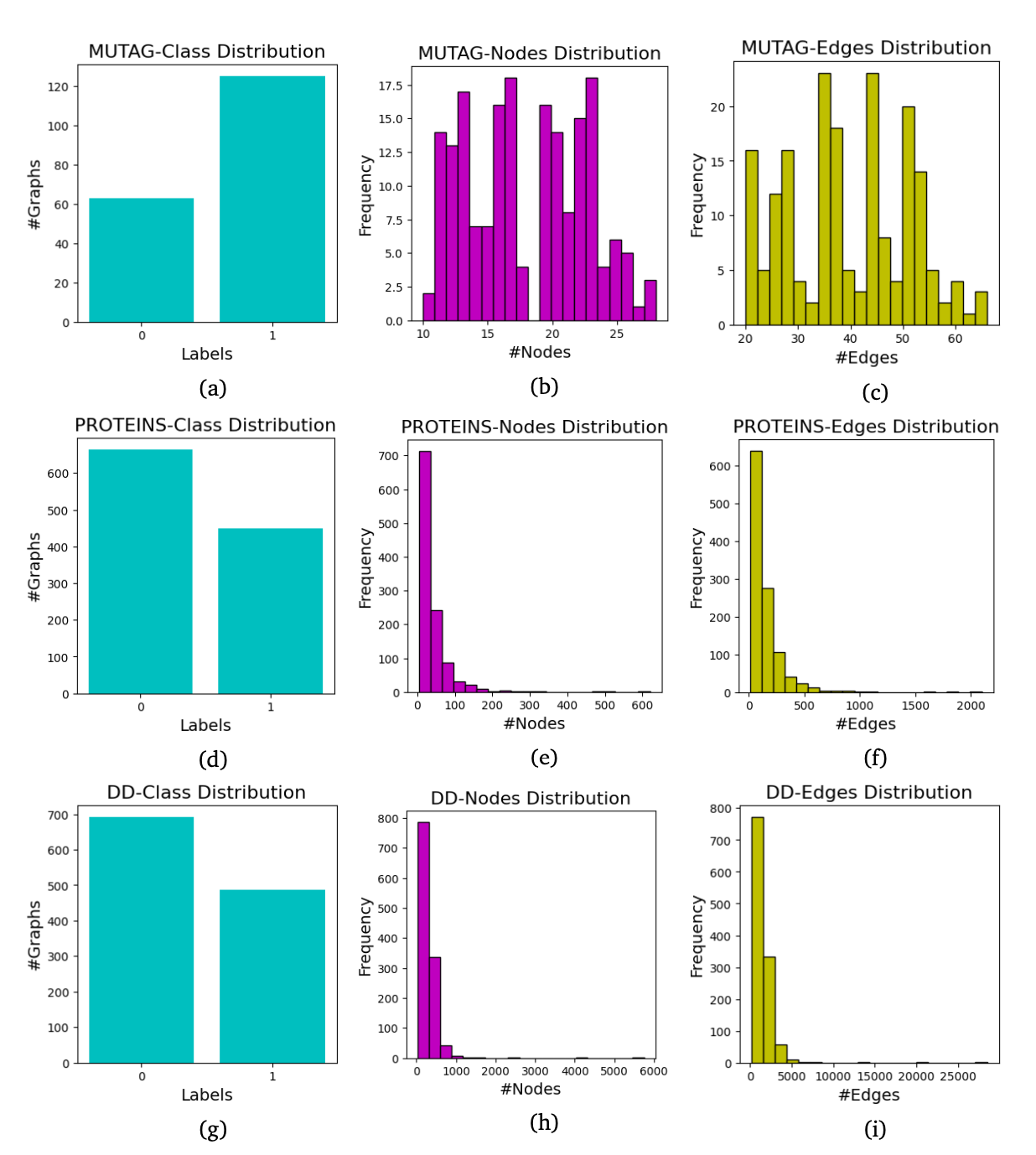}
    \caption{\textit{Data distributions of the graph classification datasets: Figures (a)-(c), (d)-(f), and (g)-(i) show the distribution of graph properties---class, nodes, and edges---in the MUTAG, PROTEINS, and DD datasets, respectively. Specifically, Figures (a), (d), and (g) illustrate class distributions; Figures (b), (e), and (h) illustrate node distributions; and Figures (c), (f), and (i) illustrate edge distributions.}}
    \label{fig:data_distribution}
\end{figure}

\renewcommand{\thefigure}{A.4}
\begin{figure}[H]
    \centering
    \includegraphics[width=1\linewidth]{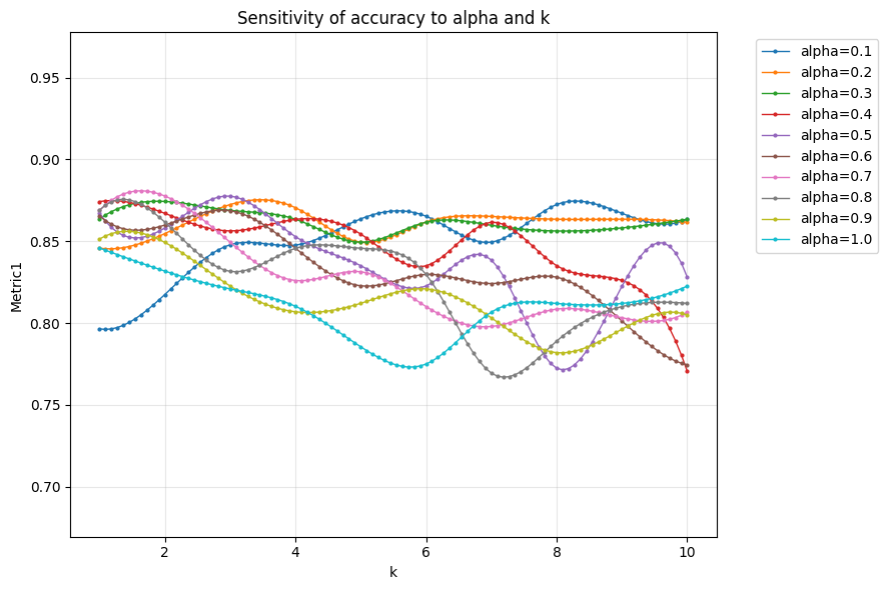}
    \caption{\textit{Extended sensitivity analysis of RGC-Net accuracy on the MUTAG dataset across different reservoir iteration counts ($k$) and leaky rates ($\alpha$). Each curve corresponds to a fixed $\alpha$ value, showing how accuracy varies as the number of reservoir iterations increases. The results indicate that RGC-Net maintains stable performance across a wide range of $\alpha$ and $k$ values, confirming the robustness of the model's dynamics and its insensitivity to small hyperparameter variations.}}
    \label{fig:appendix_sensitivity}
\end{figure}

\newpage
\section{RGC-Net is not permutation invariant}

To show that RGC-Net is not permutation invariant, we examine the effect of applying different permutations to the node order.

Let \(\mathbf{X}\) be the feature matrix and \(\mathbf{A}\) the adjacency matrix of a graph. Consider two different permutation matrices, \(\mathbf{P}_1\) and \(\mathbf{P}_2\), which rearrange the nodes in distinct ways, leading to two permuted inputs, \(\mathbf{P}_1 \mathbf{X}\) and \(\mathbf{P}_2 \mathbf{X}\).

Applying RGC-Net's transformation with \(\mathbf{P}_1 \mathbf{X}\) and \(\mathbf{P}_2 \mathbf{X}\) as inputs results in two output embeddings \(\mathbf{H}^{(k+1)}(\mathbf{P}_1 \mathbf{X})\) and \(\mathbf{H}^{(k+1)}(\mathbf{P}_2 \mathbf{X})\) after layer \(k+1\).

Due to permutation equivariance, the outputs satisfy:
\[
\mathbf{H}^{(k+1)}(\mathbf{P}_1 \mathbf{X}) = \mathbf{P}_1 \mathbf{H}^{(k+1)}(\mathbf{X}) \quad \text{and} \quad \mathbf{H}^{(k+1)}(\mathbf{P}_2 \mathbf{X}) = \mathbf{P}_2 \mathbf{H}^{(k+1)}(\mathbf{X}).
\]

However, \(\mathbf{H}^{(k+1)}(\mathbf{P}_1 \mathbf{X}) \neq \mathbf{H}^{(k+1)}(\mathbf{P}_2 \mathbf{X})\), since the embeddings depend on the specific permutation applied. This difference indicates that RGC-Net is not permutation invariant because identical embeddings would require:
\[
\mathbf{H}^{(k+1)}(\mathbf{P}_1 \mathbf{X}) = \mathbf{H}^{(k+1)}(\mathbf{X}) = \mathbf{H}^{(k+1)}(\mathbf{P}_2 \mathbf{X}),
\]
regardless of the input order. Thus, while RGC-Net is permutation equivariant, it does not satisfy permutation invariance.

\section{Datasets used for classification and generation}
\subsection*{Description of graph classification datasets}
\begin{enumerate}
    \item \textbf{MUTAG}
    Each graph in the MUTAG dataset represents a chemical compound \cite{schlichtkrull2018modeling}. In these graphs, nodes represent atoms, and edges represent chemical bonds between atoms. Each graph is labelled mutagenic (causing genetic mutations) or non-mutagenic. The edges in this graph dataset are directed and weighted.

    \item \textbf{PROTEINS}
    The PROTEINS dataset consists of graphs that represent proteins \cite{morris2020tudataset, dobson2003distinguishing}. In these graphs, nodes represent secondary protein structures (helices and sheets), and edges represent the proximity between these elements within each protein. Each protein graph is labelled as either an enzyme or a non-enzyme based on the connectivity of these structural elements. The graphs in this dataset are unweighted directed graphs.

    \item \textbf{D\&D}
    The DD dataset is derived from the same source as the PROTEINS dataset \cite{morris2020tudataset, dobson2003distinguishing}. However, instead of using secondary protein structures as nodes and their relationships as edges, the DD dataset represents graphs with amino acids as nodes and the distances between amino acids as edges. Similar to the PROTEINS dataset, the graphs in this dataset are unweighted directed graphs.
\end{enumerate}

\renewcommand{\thetable}{C.1}
\begin{table}[H]
\setlength{\tabcolsep}{5pt}
\renewcommand{\arraystretch}{1.3}
\small
\centering
\caption{Datasets used for graph classification experiments}
\label{Appendix:tab1}
\begin{tabular}{|c|c|c|c|}
\hline
\textbf{Properties} & \textbf{MUTAG} & \textbf{PROTEINS} & \textbf{D\&D} \\ \hline \hline
\textbf{General Properties} & & & \\
\#Graphs & 188 & 1113 & 1178 \\
Avg. \#Nodes per Graph & 17.93 $\pm$ 4.58 & 39.06 $\pm$ 45.76 & 284.32 $\pm$ 272.00 \\
Avg. \#Edges per Graph & 39.59 $\pm$ 11.37 & 145.63 $\pm$ 169.20 & 1431.32 $\pm$ 1387.81 \\
\#Node Features & 7 & 4 & 89 \\ \hline
\textbf{Topological Properties} & & & \\
Avg. Degree Distribution & 2.19 $\pm$ 0.11 & 3.73 $\pm$ 0.42 & 4.98 $\pm$ 0.59 \\
Avg. Clustering Coefficient & 0.00 $\pm$ 0.00 & 0.51 $\pm$ 0.23 & 0.48 $\pm$ 0.05 \\
Avg. Path Length & 3.62 $\pm$ 0.60 & 4.66 $\pm$ 2.72 & 7.95 $\pm$ 2.55 \\
Avg. Diameter & 8.22 $\pm$ 1.84 & 11.47 $\pm$ 7.85 & 19.76 $\pm$ 7.52 \\
Avg. Density & 0.14 $\pm$ 0.04 & 0.21 $\pm$ 0.20 & 0.03 $\pm$ 0.02 \\
Avg. Modularity & 0.46 $\pm$ 0.06 & 0.55 $\pm$ 0.19 & 0.785 $\pm$ 0.07 \\ \hline
\end{tabular}
\end{table}

\subsection*{Description of graph generation datasets}
\begin{enumerate}
    \item \textbf{EMCI-AD: Early Mild Cognitive Impairment and Alzheimer Disease}
    This connectomic dataset is constructed from brain imaging data from the Alzheimer's Disease Neuroimaging Initiative (ADNI) database \cite{mueller2005alzheimer}. The structural brain graphs are processed from T1-weighted MRI scans of 67 subjects' left hemispheres. Of these subjects, 32 are diagnosed with Alzheimer's Disease, and 35 with Mild Cognitive Impairment. The scans were collected twice, six months apart. The brain regions are parcellated using the Desikan--Killiany atlas \cite{desikan2006automated}, dividing the brain into 35 ROIs, resulting in brain graphs with 35 nodes.

    \item \textbf{Simulated Dataset}
    We adopted the method from Demirbilek and Rekik \cite{demirbilek2023predicting} to simulate longitudinal brain graphs of healthy adults. This simulation uses statistical data from a real connectomic dataset with $35 \times 35$ resolution, including mean connectivity values, covariance matrices, and connectivity differences over time. We generated random samples using a multivariate normal distribution based on the mean connectivity value and the covariance matrix. To simulate changes in connectivity over time, we add the connectivity value at each time point with the product of the overall connectivity difference (\( \Delta W \)) and hyperbolic tangent noise \( N_{tan} \) derived from Equation (C.1). This addition operation ensures that the graph at a later time point shows more connectivity difference from the graph in the initial time point. The connectivity weight change can be described in Equation (C.2), where \( W_{ij}^{t} \) represents the connectivity value of edge connecting node \( i \) and \( j \) at time \( t \).

\begin{equation}
N_{tan} = \tanh \left( \frac{t}{n_t} \right) = \frac{e^{\frac{t}{n_t}} - e^{-\frac{t}{n_t}}}{e^{\frac{t}{n_t}} + e^{-\frac{t}{n_t}}}
\tag{C.1}
\end{equation}

\begin{equation}
W_{ij}^{t+1} = W_{ij}^{t} + (\Delta W \cdot N_{tan})
\tag{C.2}
\end{equation}

    \item \textbf{SLIM160: Southwest University Longitudinal Imaging Multimodal}
    The SLIM (Southwest University Longitudinal Imaging Multimodal) dataset \cite{liu2017longitudinal} contains longitudinal and multi-resolution data of functional brain connectivity graphs across three time points. Each subject's brain graphs in the SLIM dataset are available in two resolutions, based on the brain atlas used during the ROI parcellation step: $160 \times 160$ resolution constructed using the Dosenbach atlas \cite{dosenbach2010prediction} and $268 \times 268$ resolution constructed using the Shen atlas \cite{shen2013groupwise}. This project only uses the brain graph data at $160\times160$ resolution.

    All brain graph data in this dataset were obtained from healthy subjects with an average age of 20 years, with approximately 55\% of the subjects being female at each time point. The average interval between the first and second data acquisitions is 304.14 days, while the interval between the second and third is 515 days. This dataset includes 580 subjects with varying acquisition times: 121 subjects have complete longitudinal data, 221 subjects have data for two time points, and 263 subjects have one time point. In this work, we only used data from the 121 subjects with complete longitudinal data. However, after initial preprocessing, we found that 12 subjects had invalid graph data with `NaN' connectivity values for all edges. As a result, only the data from 109 subjects with valid brain graphs across all three time points were used in this research.
\end{enumerate}

\renewcommand{\thetable}{C.2}
\begin{table}[H]
\setlength{\tabcolsep}{5pt}
\renewcommand{\arraystretch}{1.3}
\small
\centering
\caption{Datasets used for graph generation experiments}
\label{Appendix:tab2}  
\begin{tabular}{|c|c|c|c|}
\hline
\textbf{Properties} & \textbf{EMCI-AD} & \textbf{Simulated} & \textbf{SLIM160} \\ \hline \hline
\textbf{General Properties} & & & \\
\#Subjects & 67 & 100 & 109 \\
\#Time Points & 2 & 3 & 3 \\
\#Graphs & 134 & 300 & 327 \\
\#Nodes & 35 & 35 & 160 \\
\#Node Features & 8 & 8 & 8 \\ \hline
\textbf{Topological Properties} & & & \\
Avg. Degree Distribution & 33.66$\pm$0.00 & 34.00$\pm$0.00 & 109.24$\pm$17.07 \\
Avg. Clustering Coefficient & 0.99$\pm$0.00 & 1.00$\pm$0.00 & 0.76$\pm$0.09 \\
Avg. Path Length & 1.01$\pm$0.00 & 1.00$\pm$0.00 & 1.28$\pm$0.20 \\
Avg. Diameter & 2.00$\pm$0.00 & 1.00$\pm$0.00 & 1.96$\pm$0.29 \\
Avg. Density & 0.99$\pm$0.00 & 1.00$\pm$0.00 & 0.69$\pm$0.11 \\
Avg. Modularity & 0.00$\pm$0.00 & 0.00$\pm$0.00 & 0.12$\pm$0.04 \\ \hline
\end{tabular}
\end{table}

\end{document}